\documentclass[sigconf,screen]{acmart}
\usepackage{amsmath}
\usepackage{mathtools}
\usepackage{amsthm}
\usepackage{microtype}
\usepackage{graphicx}
\usepackage{subcaption}
\usepackage{booktabs} 
\usepackage{cleveref}
\usepackage{hyperref}
\usepackage{lineno}

\newcommand{\algname}{\textit{IMU}\xspace}

\usepackage{soul}
\usepackage{multirow}
\usepackage{adjustbox}
\usepackage[table,dvipsnames]{xcolor}
\definecolor{grey}{RGB}{128,138,135}
\definecolor{darkgrey}{RGB}{96,96,96}
\usepackage{pifont}
\newcommand{\cmark}{\textcolor{green!60!black}{\ding{51}}} 
\newcommand{\xmark}{\textcolor{red}{\ding{55}}}             
\def\with{\textbf{w/}\xspace}
\def\without{\textbf{w/o}\xspace}

\usepackage{algorithm}
\usepackage{algorithmic}

\usepackage{empheq}
\newcommand*\widefbox[1]{\fbox{\hspace{0.5em}#1\hspace{0.2em}}}

\usepackage{soul}

\RequirePackage{xspace}
\makeatletter
\DeclareRobustCommand\onedot{\futurelet\@let@token\@onedot}
\def\@onedot{\ifx\@let@token.\else.\null\fi\xspace}

\def\eg{\emph{e.g}\onedot} 

\def\ie{\emph{i.e}\onedot}

\def\vs{\emph{vs}\onedot}
\def\wrt{w.r.t\onedot}

\makeatother

\usepackage{amsthm}

\usepackage{amsmath,amsfonts,bm}

\def\eqref#1{equation~\ref{#1}}

\def\1{\bm{1}}

\def\vone{{\bm{1}}}

\def\vtheta{{\bm{\theta}}}
\def\va{{\bm{a}}}

\def\vg{{\bm{g}}}

\def\vp{{\bm{p}}}

\def\vs{{\bm{s}}}

\def\vx{{\bm{x}}}
\def\vy{{\bm{y}}}
\def\vz{{\bm{z}}}

\def\mA{{\bm{A}}}

\def\mE{{\bm{E}}}

\def\mH{{\bm{H}}}

\def\mJ{{\bm{J}}}
\def\mK{{\bm{K}}}

\def\mS{{\bm{S}}}

\def\mW{{\bm{W}}}
\def\mX{{\bm{X}}}

\DeclareMathAlphabet{\mathsfit}{\encodingdefault}{\sfdefault}{m}{sl}
\SetMathAlphabet{\mathsfit}{bold}{\encodingdefault}{\sfdefault}{bx}{n}

\def\gD{{\mathcal{D}}}

\def\gI{{\mathcal{I}}}

\def\gL{{\mathcal{L}}}

\def\gO{{\mathcal{O}}}

\newcommand{\E}{\mathbb{E}}

\DeclareMathOperator*{\argmin}{arg\,min}

\theoremstyle{plain}
\newtheorem{theorem}{Theorem}[section]

\theoremstyle{definition}

\usepackage{amsthm}
\newtheorem{Assumption}{Assumption}
\theoremstyle{remark}
\newtheorem{remark}[theorem]{Remark}
\AtBeginDocument{%
  }

\setcopyright{acmlicensed}
\copyrightyear{2026}
\acmYear{2026}
\acmDOI{XXXXXXX.XXXXXXX}
\acmConference[Preprint]{}{}{}
\acmISBN{978-1-4503-XXXX-X/2026/06}

\renewcommand\footnotetextcopyrightpermission[1]{}

\begin{document}

\title{IMU: Influence-guided Machine Unlearning}


\author{Xindi Fan}
\email{xindi.fan@stu.hit.edu.cn}
\affiliation{%
  \institution{Harbin Institute of Technology}
  \city{Harbin}
  \country{China}
}
\author{Jing Wu}\authornote{Corresponding author}
\email{jing.wu2@monash.edu}
\affiliation{%
  \institution{Monash University}
  \city{Melbourne}
  \state{VIC}
  \country{Australia}
}
\author{Mingyi Zhou}
\email{zhoumingyi@buaa.edu.cn.}
\affiliation{%
  \institution{Beihang University}
  \city{Beijing}
  \country{China}
}
\author{Pengwei Liang}
\email{penliang@cityu.edu.hk}
\affiliation{%
  \institution{City University of Hong Kong}
  \city{Hong Kong}
  \country{SAR}
}
\author{Mehrtash Harandi}
\email{mehrtash.harandi@monash.edu}
\affiliation{%
  \institution{Monash University}
  \city{Melbourne}
  \state{VIC}
  \country{Australia}
}
\author{Dinh Phung}
\email{dinh.phung@monash.edu}
\affiliation{%
  \institution{Monash University}
  \city{Melbourne}
  \state{VIC}
  \country{Australia}
}


\renewcommand{\shortauthors}{Xindi Fan et al.}
\begin{abstract}
Machine Unlearning (MU) aims to selectively erase the influence of specific data points from pretrained models. However, most existing MU methods rely on the retain set to preserve model utility, which is often impractical due to privacy restrictions and storage constraints.
While several retain-data-free methods attempt to bypass this using geometric feature shifts or auxiliary statistics, they typically treat forgetting samples uniformly, overlooking their heterogeneous contributions.
To address this, we propose \ul{I}nfluence-guided \ul{M}achine \ul{U}nlearning (IMU), a principled method that conducts MU using only the forget set. 
Departing from uniform Gradient Ascent (GA) or implicit weighting mechanisms, IMU leverages influence functions as an explicit priority signal to allocate unlearning strength.
To circumvent the prohibitive cost of full-model Hessian inversion, we introduce a theoretically grounded classifier-level influence approximation.
This efficient design allows IMU to dynamically reweight unlearning updates, aggressively targeting samples that most strongly support the forgetting objective while minimizing unnecessary perturbation to retained knowledge.
Extensive experiments across vision and language tasks show that IMU achieves highly competitive results.
Compared to standard uniform GA, IMU maintains identical unlearning depth while enhancing model utility by an average of 30\%, effectively overcoming the inherent utility-forgetting trade-off.
\end{abstract}





\maketitle

\section{Introduction}
\label{sec:intro}

\begin{figure*}[tb]
    \centering
    \includegraphics[width=0.86\textwidth]{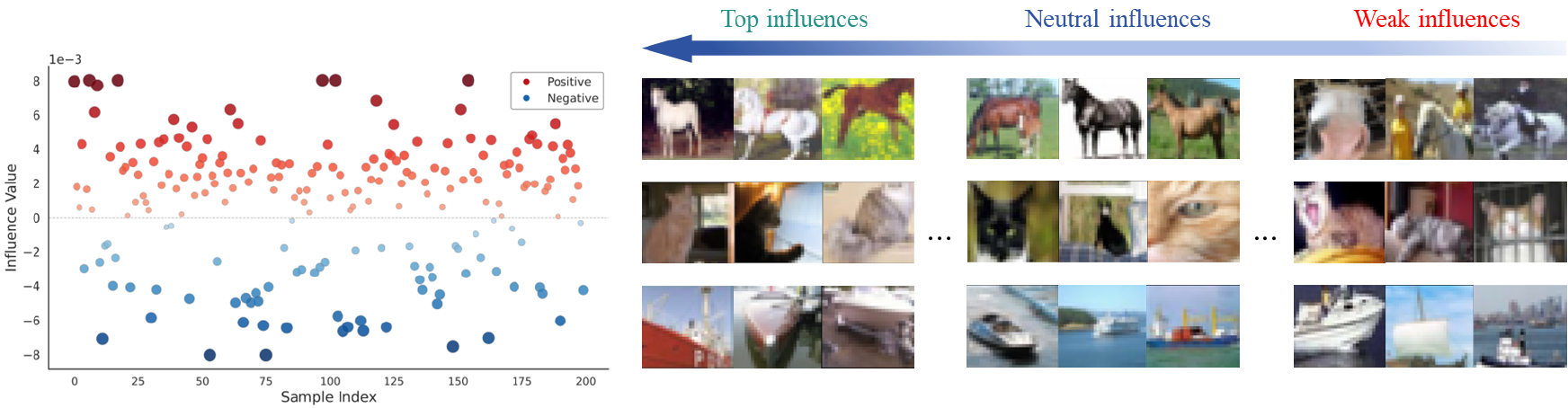}
    \caption{
    Visualizing the heterogeneous influence of the forgetting data.
    (Left) The scatter plot reveals a wide variance in per-sample influence values, confirming that individual data points contribute non-uniformly to the learned model. (Right) Example images categorized by their influence magnitudes.
    }
    \label{fig:sample_if}
\end{figure*}

Deep learning (DL) models are widely deployed in various applications, but their vulnerability to adversarial attacks, such as membership inference attacks~\cite{rezaei2021difficulty} and model inversion attacks~\cite{zhu2019deep,geiping2020inverting,balunovic2021bayesian}, raises significant concerns about privacy leakage.
In response, legislation such as GDPR~\cite{voigt2017eu} grants users the \textit{right to be forgotten}, compelling models to remove data points upon request. 
Retraining from scratch is often computationally prohibitive for modern large-scale models, motivating the development of machine unlearning (MU), \ie, a paradigm aimed at efficiently removing the effect of the forgetting data while preserving utility on retain data.

Current MU methods typically require fine-tuning on the retain data, which are part of the original training data. 
However, this retain data may be entirely unavailable due to privacy restrictions, ownership, or storage constraints in many real-world deployments. 
Recently, a few studies have sought to overcome this limitation by developing retain-data-free MU methods~\cite{cha2024learning,bonato2024retain,foster2024information}.
While these methods show promising unlearning capabilities, they often operate under restrictive assumptions (\eg, requiring access to auxiliary data or stored dataset statistics).

More importantly, a key challenge that remains underexplored is the heterogeneous contribution of individual forgetting data points.
In practice, forgetting data points exhibit highly non-uniform influence on the learned model: a small subset may contribute dominantly to the model's fit on the forget set $\gD_f$, while others provide marginal or redundant contributions, and may even introduce unnecessary drift to utility-relevant parameters when treated uniformly during unlearning.
\Cref{fig:sample_if} illustrates this heterogeneity when conducting class-wise and sample-wise unlearning in CIFAR-10, where per-sample influence values span a wide range.
Samples with high removal influence (\ie, whose removal would significantly raise the loss on $\gD_f$) represent data points that the model has fitted most deeply; they therefore require the strongest unlearning effort to effectively erase.
In contrast, low-influence samples contribute little to the model's knowledge of $\gD_f$, and aggressively unlearning them provides diminishing returns while risking unnecessary perturbation to retained knowledge.

To address this heterogeneous nature, we propose Influence-guided Machine Unlearning (\algname), a retain-data-free framework that dynamically allocates unlearning strength based on each sample's true contribution. 
Instead of relying on standard uniform Gradient Ascent (GA), which blindly updates parameters and risks catastrophic utility collapse, we leverage influence functions~\cite{koh2017understanding} as an explicit \emph{priority signal}. 
By quantifying how removing a specific data point affects the forgetting objective, \algname directs the limited unlearning budget to where it is most needed. It aggressively erases high-influence samples while safeguarding the model from unnecessary drift caused by less informative ones.

However, directly applying influence functions to deep neural networks presents significant practical challenges. 
Theoretical influence estimation relies on convex loss landscapes, and computing the required inverse Hessian-vector products is computationally prohibitive for modern models with millions of parameters~\cite{basu2020influence,mehta2022deep}. 
To overcome these computational bottlenecks, we introduce a theoretically grounded classifier-level influence approximation. 
By viewing the network as a fixed feature extractor followed by a shallow prediction head, we restrict the Hessian computation exclusively to the classifier. 
This formulation not only yields a convex subproblem for reliable estimation but also drastically reduces the computational overhead.
Ultimately, \algname provides a principled, highly efficient approach to retain-data-free MU. Experimental results demonstrate that our proposed algorithm effectively removes the influence of forgetting data while preserving model utility, consistently outperforming existing retain-data-free baselines.

Our contributions are as follows:
\begin{itemize}

    \item We propose \algname, a principled retain-data-free MU method that explicitly addresses the heterogeneous contributions of forgetting data. By utilizing influence scores as a priority signal, \algname dynamically allocates unlearning strength, \ie, aggressively targeting high-influence samples while safeguarding retained knowledge from less informative ones.
    
    \item To enable the efficient use of influence functions within this framework, we introduce a theoretically grounded classifier-level influence approximation. This design successfully circumvents the intractable cost of full-model Hessian inversion on deep non-convex models, maintaining high estimation fidelity with minimal overhead.

    \item Extensive experiments across vision and language tasks show that \algname consistently outperforms existing retain-data-free MU methods, achieving a comparable balance between unlearning efficacy and model utility preservation.

\end{itemize}

\section{Related Work}

\subsection{Machine Unlearning}
MU algorithms typically fall into two categories: exact MU methods~\cite{ginart2019making,cao2015towards, bourtoule2021machine} that use data partitioning strategies or only focus on traditional machine learning models like $k$-means clustering~\cite{ginart2019making}, and approximate MU methods that trade perfect deletion for computational efficiency. 
The latter include parameter perturbation~\cite{golatkar2020eternal,chen2023boundary,kumari2023ablating,foster2024fast}, incremental strategies~\cite{heng2023continual,kurmanji2023towards,lyu2024one,bui2024removing}, and sparsity or data-attribution techniques~\cite{ko2024boosting,lin2024gdr,spartalis2025lotus,alberti2025data,wu2025erasing,foster2024lossfreemachineunlearning,goel2023adversarialevaluationsinexactmachine}.
So far, these MU methods have demonstrated effectiveness through two key metrics: (1) degraded performance on the forgetting data, and (2) maintained utility on retain and unseen set.

Gradient ascent (GA)-based MU~\cite{wu2020deltagrad,gandikota2023erasing,jang2022knowledge} perform unlearning by moving model parameters away from the forgetting data through gradient ascent.
While simple to implement, in practice, these methods require control mechanisms to prevent damaging the model’s utility~\cite{jia2023model,fan2024simplicity}.
Fisher unlearning~\cite{golatkar2020eternal,golatkar2020forgetting,golatkar2021mixed} deem the unlearned model and the retrained model are close to each other in parameter space, and formulate Fisher Forgetting, which induces unlearning by injecting noise into the parameters proportional to their relative importance in the forgetting data compared to the retain set.
Influence function-based methods estimate the impact of individual data points to perform approximate unlearning.
\citet{guo2019certified} propose a one‐step Newton‐update procedure based on the influence function to excise the effect of specific data points from a pretrained model. Subsequent works~\cite{mehta2022deep,liu2022right,jia2023model} focus on ways for more efficient computation over the estimation.
Two-stage methods~\cite{jia2023model,fan2023salun} have emerged as a practical alternative, first erasing information related to the forgetting data, then fine-tuning on retain data. These methods avoid expensive computations while achieving strong empirical results, particularly when combined with sparsity techniques.

\noindent
\textbf{Retain-data-free MU.}
While the aforementioned methods show strong empirical results, they rely on access to the retain set, presenting challenges under strict privacy or storage constraints.
Recent studies~\cite{cha2024learning,bonato2024retain,foster2024information} have pioneered retain-data-free methods to address this.
\citet{cha2024learning} propose instance-wise unlearning by generating and fine-tuning on adversarial examples \wrt the forgetting data. \citet{bonato2024retain} leverages Mahalanobis distance to shift forgetting representations toward incorrect classes, followed by knowledge distillation with auxiliary images for preserving model utility. In the large language model (LLM) domain, Negative Preference Optimization (NPO)~\cite{zhang2024negative,fan2024simplicity} treats forgetting data as negative examples within a preference optimization framework.

Despite their promising unlearning capabilities, a critical aspect that remains underexplored across these methods is the heterogeneous contributions of individual forgetting data points.
Most existing approaches apply a uniform unlearning objective across the entire forget set. While recent frameworks like NPO implicitly introduce instance-level weighting through preference optimization, they scale updates based on output-level prediction confidence (\eg, log-likelihoods) rather than a direct measure of a sample's parametric contribution.
Consequently, these methods risk over-unlearning low-influence samples or under-unlearning critical ones. 
Complementing these pioneering efforts, our work explicitly tackles this heterogeneity by introducing a theoretically grounded framework that dynamically allocates unlearning budget based on the precise parametric influence of each data point.

\subsection{Influence Function in Deep Learning}
Influence functions~\cite{hampel1974influence,koh2017understanding,pruthi2020estimating} assess how infinitesimal changes to training data weights affect performance on a validation set, by analyzing their impact on a target evaluation metric.
For linear models, influence functions are well-defined thanks to the convexity of the loss landscape, but in deep learning, where loss landscapes are typically non-convex, the behavior of influence functions remains poorly understood~\cite{basu2020influence}. 
\citet{basu2020influence} presents a comprehensive empirical study on the reliability of influence functions in deep neural networks (DNNs).
The findings reveal that factors such as network architecture, depth, width, parameterization, and regularization significantly impact influence estimation accuracy.
Overall, the results demonstrate that influence functions in DNNs frequently struggle to accurately predict the effects of retraining, leading to the conclusion that influence estimates are often brittle and unreliable.

Although recent works~\cite{bae2022if,epifano2023revisiting} have defended their applicability in DNNs, extending them remains non-trivial.
Existing strategies include replacing non-convex embeddings with linear ones~\cite{li2022achieving,chhabra2024data}, adding damping terms to ensure positive-definite Hessians~\cite{koh2017understanding,han2020explaining}, or deriving second-order formulations~\cite{basu2020second,alaa2020discriminative}.
Furthermore, computing the required inverse Hessian-vector products is computationally prohibitive for modern large-scale models~\cite{mehta2022deep}. To bypass these limitations, we introduce a theoretically grounded classifier-level approximation, enabling the robust and efficient use of influence functions for MU without requiring full-model Hessian inversion.

\section{Methodology}
In this section, we propose \algname, a retain-data-free unlearning framework that scrubs data from a model by dynamically eliminating its influence from the model.
We consider the unlearning setting where, upon a deletion request, the unlearning procedure is only allowed to access $\gD_f$ and the current model, while $\gD_r$ is unavailable due to privacy, storage, or ownership constraints.
Throughout the paper, we denote scalars and vectors/matrices by lowercase and bold symbols, respectively (\eg, $a$, $\va$, and $\mA$).

\subsection{Preliminaries}

\textbf{Notation.}
Let $f_\vtheta: \mathcal{X} \to \mathcal{Y}$ be a model with parameters $\vtheta$, classifying inputs $\vx \in \mathcal{X}$ 
to labels $\vy$ in the label space $\mathcal{Y}$.
Let $\gD_{\text{train}} = \{(\vx_i, \vy_i)\}_{i=1}^N$ denote our full training set.
We partition $\gD_{\text{train}}$ into two disjoint subsets: the forgetting data $\gD_f\subset \gD_{\text{train}}$, whose samples are required to be removed from the model, and the retain data $\gD_r = \gD_{\text{train}}\setminus \gD_f$, whose knowledge we wish to preserve.
We decompose $f_\vtheta$ into two modules: a feature extractor $\phi_{\vtheta_{e}} \colon \mathbb{R}^n\to\mathbb{R}^d$, comprising all layers up to but excluding the final classification stage, and a classifier $h_{\vtheta_{c}} \colon \mathbb{R}^d\to\mathbb{R}^C$, corresponding to the last fully connected layer.
%
Our goal is to adjust $f_\vtheta$ so that it effectively erases the information associated with $\gD_f$ while maintaining performance on $\gD_r$ and the unseen set $\gD_t$.

\noindent
\textbf{Influence function.}
Refer to \cite{koh2017understanding}, we denote the empirical risk minimizer of the loss over $\gD_{\text{train}}$ as $\vtheta^\ast \triangleq \argmin_{\vtheta}\frac{1}{N}\sum_{i=1}^N \ell(f_\vtheta(\vx_i))$.
If upweighting the training data point $\vx$ by some small $\epsilon$, the new optimal parameter is $\hat{\vtheta}_{\epsilon, \vx} \triangleq \argmin_{\vtheta}\frac{1}{N}\sum_{i=1}^{N} \ell(f_\vtheta(\vx_i)) + \epsilon \ell(f_\vtheta(\vx))$.
The change in model parameters can then be approximated by
\begin{align}
    \gI_{\texttt{up}, \vtheta^\ast}(\vx) \triangleq \left.\frac{d \hat{\vtheta}_{\epsilon,\vx}}{d \epsilon} \right|_{\epsilon=0}= -\mH_{\vtheta^\ast}^{-1} \nabla_{\vtheta} \ell \big(f_{\vtheta^\ast}(\vx) \big),
\end{align}
where $\mH_{\vtheta^\ast}=\frac{1}{N}\sum_{i=1}^{N}\nabla^2_{\vtheta} \ell \big(f_{\vtheta^\ast}(\vx_i)\big)$.
Consequently, the influence on the loss for $\gD_t$ when upweighting $\vx$ would be:
\begin{align}
    \gI_{\texttt{up}, \ell}(\vx, \gD_t) &\triangleq \left.\frac{d \ell\left(f_{\hat{\vtheta}_{\epsilon,\vx}}(\gD_t)\right) }{d \epsilon} \right|_{\epsilon=0} \notag \\
    =- &\nabla_{\vtheta} \ell \big(f_{\vtheta^\ast}(\gD_t)\big)^\top \mH_{\vtheta^\ast}^{-1} \nabla_{\vtheta} \ell \big(f_{\vtheta^\ast}(\vx) \big).
\end{align}
Proofs can be found in ~\cite{koh2017understanding}.
However, this classical derivation assumes sufficient smoothness and local invertibility of the Hessian, which may not strictly hold for highly non-convex deep networks.
Besides, for DNNs, computing the exact inverse of the Hessian matrix is computationally expensive~\cite{mehta2022deep}.

\subsection{IMU}
As discussed in \textsection~\ref{sec:intro}, a critical aspect that remains underexplored across retain-data-free MU methods is the heterogeneous contributions of individual forgetting data points.
Treating all forgetting samples uniformly (\eg, GA) ignores the fact that certain data points contribute disproportionately to the model's learned representations. 
To achieve precise and effective unlearning, it is crucial to dynamically allocate the unlearning budget based on each sample's actual parametric contribution.
While influence functions provide a principled mathematical framework to quantify this sample-wise contribution, directly applying them to DNNs is practically prohibited by the computational cost and instability of high-dimensional Hessians. To bridge this gap, \algname proposes a robust, classifier-level influence estimation strategy to efficiently guide the MU process.

\noindent
\textbf{Influence estimation.}
According to the manifold hypothesis, high-dimensional data tends to lie on lower-dimensional manifolds~\cite{bengio2013representation}.
Hence, rather than estimating influence on the full deep network, which would require inverting a high-dimensional, potentially ill-conditioned Hessian, we instead compute influence at the level of learned representations.
Specifically, we treat the feature representation $\vz=\phi(\vx)$ as the input and estimate the influence of $\vx$ \wrt the classifier $h(\cdot)$.
We'll formally justify this approximation in \textsection~\ref{subsec:justify}.

Removing $\vx$ corresponds to setting $\epsilon=-1/N$, therefore, we estimate the parameter change under removal as
\begin{align}
    \Delta \vtheta^\ast_{c,-\vx} \approx - \frac{1}{N}\gI^c_{\texttt{up}, \vtheta^\ast_c}(\vx) = \frac{1}{N}\mH_{\vtheta_c^\ast}^{-1} \nabla_{\vtheta_c} \ell \big( h_{\vtheta_c^\ast} (\vz) \big),
\end{align}
where the latent representation $\vz = \phi_{\vtheta_e^\ast}(\vx)$.
Accordingly, the change in the loss over $\gD_f$ induced by removing $\vx$ is
\begin{align}
\label{eqn:if}
    \Delta_{-\vx} &\ell(\gD_f) \approx (-1/N)\; \gI^c_{\texttt{up}, \ell}(\vx, \gD_f) \notag \\
    = &(1/N)\; \nabla_{\vtheta_c} \ell\big(h_{\vtheta_c^\ast}(\mathbf{Z}_f)\big)^\top \mH_{\vtheta_c^\ast}^{-1} \nabla_{\vtheta_c} \ell\big(h_{\vtheta_c^\ast}(\vz)\big),
\end{align}
where $\mathbf{Z}_f = \phi_{\vtheta_e^\ast}(\gD_f)$.
Then $\Delta_{-\vx} \ell(\gD_f) > 0$ indicates that $\vx$ is helpful for learning $\gD_f$, whereas $\Delta_{-\vx} \ell(\gD_f) < 0$ indicates that $\vx$ is harmful to $\gD_f$.

Given the access constraint ($\gD_r$ is unavailable), computing the exact full-data Hessian is infeasible. Therefore, we use $\mH^f_{\vtheta_c^\ast} \triangleq \nabla^2_{\vtheta_c}\ell(h_{\vtheta_c^\ast}(\mathbf{Z}_f))$, which captures the local curvature of the forgetting objective around $\vtheta_c^\ast$.
While this omits the global curvature contributed by $\gD_r$, it explicitly captures the local geometric structure of the forgetting subspace.
This yields a targeted influence estimate that is sufficient for our purpose of weighting $\vx \sim \gD_f$.


\noindent
\textbf{Influence-guided MU.}
To dynamically allocate unlearning budgets based on sample influence, we integrate these influence scores into the MU process and define the following influence-guided unlearning loss:
\begin{align}
    \label{eqn:imu_loss}
    \gL_{\texttt{IMU}}(\gD_f; \vtheta) &= -\sum_{\vx_i \in \gD_f^+} \hat{w}_i \cdot \ell \big( f_\vtheta(\vx_i) \big),
\end{align}
where $\hat{w}_i$ is the normalized influence scores for sample $\vx_i$ and is computed as:
\begin{align}
\label{eqn:w}
    \hat{w}_i = \frac{\psi(w_i)} {\sum_{\vx_j\in\gD_f^+} \psi(w_j)}, w_i = \max\!\big(0,\, \Delta_{-\vx_i} \ell(\gD_f) \big).
\end{align}
Since our objective is to remove knowledge about $\gD_f$, we prioritize the helpful data points ($\Delta_{-\vx} \ell(\gD_f)>0$) whose removal would naturally increase the loss and restrict the weighting to the subset $\gD_f^+ \triangleq \{\vx_i \in \gD_f \mid \Delta_{-\vx} \ell(\gD_f)>0\}$.

\begin{algorithm}[tb]
\caption{Influence-guided Machine Unlearning (\algname).}
\label{alg: pseudocode}
\begin{algorithmic}[1]
    \REQUIRE Model $f$ parameterized by $\vtheta$, consist of a feature extractor $\phi$ with parameters $\vtheta_e$ and a classifier $h$ parameterized by $\vtheta_c$, forgetting data $\gD_f$.
    \ENSURE Parameters $\vtheta^*$ for the scrubbed model.
    \STATE $\vtheta^{0}=\vtheta$, learning rate $\eta$, number of iterations $T$.
    \FOR{iteration $t$ in $T$}
        \STATE Compute $\mH_{\vtheta_c}=\E_{\vz_i \in \gD_f} \left[\nabla^2_{\vtheta_c} \ell \left( h_{\vtheta_c}\left(\vz_i\right) \right) \right]$ where $\vz_i = \phi_{\vtheta_e}(\vx_i), \forall \vx_i \sim \gD_f$. \\
        {\color{gray}/* Estimate the parameter change under removal */}
        \STATE Estimate $\Delta_{-\vx} \ell(\gD_f)$: \\
        \quad \quad $(1/N)\;\nabla_{\vtheta_c} \ell\big( h_{\vtheta_c}(\mathbf{Z}_f)\big)^\top \mH_{\vtheta_c}^{-1} \nabla_{\vtheta_c} \ell\big( h_{\vtheta_c} (\vz)\big)$. \\
        {\color{gray}/* Process with samples that are helpful to $\gD_f$ */}
        \STATE Set $w_i = \max\!\big(0,\, \Delta_{-\vx_i} \ell(\gD_f) \big), \hat{w}_i = \frac{\psi(w_i)} {\sum_{\in\gD_f^+} \psi(w_j)},$ \\
        where $\gD_f^+ \triangleq \{\vx_i \in \gD_f \mid \Delta_{-\vx} \ell(\gD_f)>0\}$. \\
        {\color{gray}/* Dynamically allocates unlearning strength */}
        \STATE Compute the influence-guided loss $\gL_{\texttt{IMU}}(\gD_f; \vtheta^{(t)})$: \\
        \quad \quad $-\sum_{\vx_i \in \gD_f^+} \hat{w_i} \cdot \ell \left( f_{\vtheta} \left(\vx_i \right)\right) $.
        \STATE Updating: $\vtheta_c^{(t+1)} = \vtheta_c^{(t)} - \eta \nabla_{\vtheta_c^{t}} \gL_{\texttt{IMU}}(\gD_f; \vtheta^{(t)})$.
    \ENDFOR
    \STATE \textbf{return} $\vtheta^{(T)}$
\end{algorithmic}
\end{algorithm}
The overall procedure of \algname is summarized in \Cref{alg: pseudocode}.
\algname uses these estimated influence scores to dynamically allocate unlearning budgets across the forgetting set.
Samples with higher influence scores, which indicate a stronger contribution to the memorization of $\gD_f$, receive proportionally more aggressive updates. 
This sample-specific attention enables targeted and efficient forgetting while preserving the model's performance on retain and unseen data.
To theoretically ground our approach, we dedicate the following subsection to formally justifying our core approximation: estimating the influence at the classifier level.

\subsection{Validity of Classifier-level Approximation}
\label{subsec:justify}

\noindent
\textbf{Theoretical analysis.}
A key design choice in \algname is to estimate influence exclusively at the classifier layer.
This choice is motivated by both practical and statistical considerations: full-model influence estimation in DNNs is computationally expensive due to high-dimensional Hessians, and is often unstable due to the non-convexity. We therefore ask whether classifier-level estimation can serve as a reliable surrogate.
Motivated by extensive empirical findings in transfer learning, lower-layer representations in well-converged networks tend to be general-purpose and stable, while task-specific adaptation is primarily concentrated in higher layers or the classifier head~\cite{sagun2018empirical, yosinski2014transferablefeaturesdeepneural, donahue2013decafdeepconvolutionalactivation,howard2018universallanguagemodelfinetuning,kornblith2019betterimagenetmodelstransfer}. 
Building on these established phenomena, we formulate \Cref{assumption} as a local analytical abstraction to theoretically justify our approach.

%
\begin{Assumption}[Feature Stability and Weak Coupling]
\label{assumption}
Let the model parameters be partitioned as $\vtheta = [\vtheta_e^\top, \vtheta_c^\top]^\top$, where $\vtheta_e$ and $\vtheta_c$ denote the parameters of the feature extractor and the classifier, respectively. 
Let $\gL_f(\vtheta)\triangleq \ell(f_\vtheta(\gD_f))$, and let its Hessian at the converged parameter $\vtheta^*$ be block-partitioned accordingly as
\begin{align}
\mH^f_{\vtheta^*}
=
\begin{bmatrix}
\mH_{ee} & \mH_{ec} \\
\mH_{ce} & \mH_{cc}
\end{bmatrix},
\end{align}
Assuming the presence of convex regularization or Hessian damping standard in influence function estimation~\cite{koh2017understanding}, we assume that:
\begin{enumerate}
    \item $\mH_{ee}$ and $\mH_{cc}$ are strictly positive definite and invertible, such that $\|\mH_{ee}^{-1}\|$ and $\|\mH_{cc}^{-1}\|$ are $\gO(1)$;

    \item the feature-classifier Hessian coupling is weak, satisfying $\|\mH_{ce}\mH_{ee}^{-1}\mH_{ec}\| \ll 1;$

    \item the feature-level gradients of both the forgetting objective and an individual sample are sufficiently small, \\ \ie, $\|\nabla_{\vtheta_e}\gL_f(\vtheta^*) \ll 1\|$  and $\|\nabla_{\vtheta_e}\ell(f_{\vtheta^*}(\vx))\| \ll 1.$ 
\end{enumerate}
\end{Assumption}

\begin{remark}
The aforementioned conditions are standard analytical abstractions in both second-order optimization and influence function literature. Specifically, the strictly positive definite condition via Hessian damping (Condition 1) is a ubiquitous prerequisite for ensuring tractable influence estimation in deep networks~\citep{koh2017understanding}. Furthermore, the weak feature-classifier coupling (Condition 2) formally encapsulates the block-diagonal Hessian approximations fundamentally relied upon by scalable second-order methods, such as K-FAC~\cite{martens2015optimizing, donahue2013decafdeepconvolutionalactivation}. Finally, the small gradient assumption (Condition 3) naturally follows from the standard first-order optimality condition near a converged local minimum~\cite{nocedal1999numerical, kornblith2019betterimagenetmodelstransfer}.
\end{remark}

Under \Cref{assumption}, we can establish the following theorem, which directly matches the removal score used in \Cref{eqn:if}. The proof is deferred to \textsection~\ref{classifier-level proof} in the appendix.
%
\begin{theorem}[Approximation Error]
\label{theorem:cls_influence_approximation}
Denote the full-model and classifier-level removal scores for a sample $\vx$ as $\Delta^{\mathrm{full}}_{-\vx}\ell(\gD_f)$ and $\Delta^{c}_{-\vx}\ell(\gD_f)$, which is defined respectively as:
\begin{align}
\Delta^{\mathrm{full}}_{-\vx}\ell(\gD_f)
&\triangleq
\frac{1}{N}
\nabla_{\vtheta}\gL_f(\vtheta^*)^\top
(\mH^f_{\vtheta^*})^{-1}
\nabla_{\vtheta}\ell(f_{\vtheta^*}(\vx)), \\
\Delta^{c}_{-\vx}\ell(\gD_f)
&\triangleq
\frac{1}{N}
\nabla_{\vtheta_c}\gL_f(\vtheta^*)^\top
\mH_{cc}^{-1}
\nabla_{\vtheta_c}\ell(f_{\vtheta^*}(\vx)).
\end{align}
Under \Cref{assumption}, the absolute approximation error is bounded by:
\begin{align}
&\Big|\Delta^{\mathrm{full}}_{-\vx}\ell(\gD_f) - \Delta^{c}_{-\vx}\ell(\gD_f) \Big|  \notag\\
=
\mathcal{O}\!\Bigg(&\frac{1}{N}
\Big(
\|\nabla_{\vtheta_e}\gL_f(\vtheta^*)\|\,\|\mH_{ee}^{-1}\mH_{ec}\|
+ \|\mH_{ce}\mH_{ee}^{-1}\|\,\|\nabla_{\vtheta_e}\ell(f_{\vtheta^*}(\vx))\| \notag\\
+
\|&\nabla_{\vtheta_e}\gL_f(\vtheta^*)\|\,\|\nabla_{\vtheta_e}\ell(f_{\vtheta^*}(\vx))\|
+ \|\mH_{ce}\mH_{ee}^{-1}\mH_{ec}\|
\Big)
\Bigg) \ll 1.
\end{align}
\end{theorem}
Note that every component constituting the right-hand side of \Cref{theorem:cls_influence_approximation} is small under \Cref{assumption}, consequently, the approximation gap satisfies $|\Delta^{\mathrm{full}}_{-\vx}\ell(\gD_f) - \Delta^{c}_{-\vx}\ell(\gD_f)| \ll 1$, analytically certifying that the classifier-level score in \Cref{eqn:if} faithfully preserves the dominant signal of the full-model removal score.

\begin{figure}[tb]
    \centering
    \includegraphics[width=0.30\textwidth]{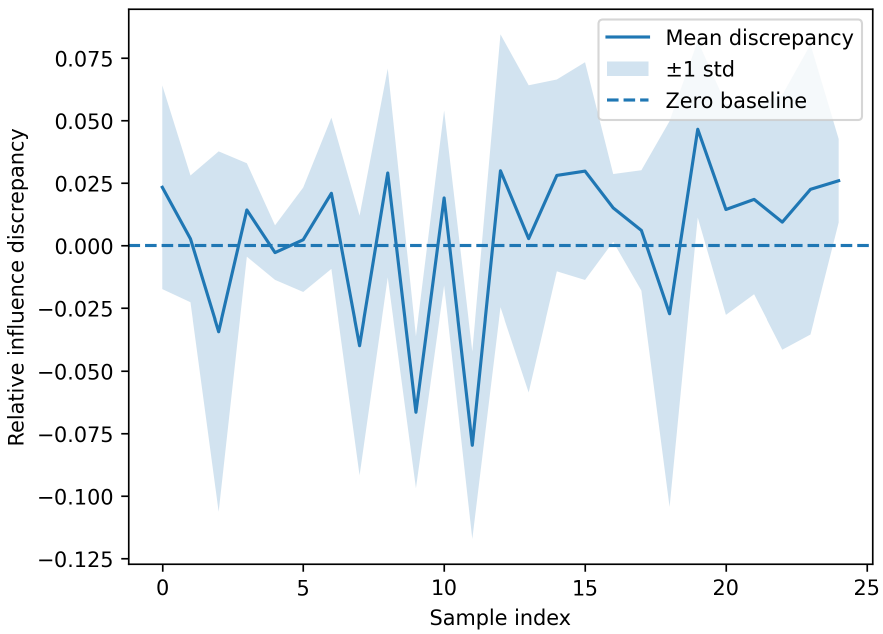}
    \caption{
    Relative discrepancy between our computationally efficient classifier-level estimation and the exact full-model influence across individual samples.
    }
    \label{fig:influence_alignment}
\end{figure}
\noindent
\textbf{Empirical validation.}
We further empirically evaluate the classifier-level approximation.
Because \algname utilizes influence scores predominantly for sample ranking rather than exact magnitude estimation, the critical criterion is whether the computationally efficient classifier-level score preserves the relative ordering induced by the full-model influence.
As visually depicted in \Cref{fig:influence_alignment}, the two estimations exhibit alignment across samples.
Furthermore, when evaluated across five independent random seeds, the classifier-level approximation achieves a striking mean Spearman's rank correlation coefficient of $0.992$ and a Kendall's coefficient of concordance of $0.984$.
These rank-correlation metrics empirically validate that our localized approximation is not only theoretically sound but also highly effective and robust to stochastic training variations.

\begin{remark}
We evaluate influence estimation across various computational paradigms in \textsection~\ref{subsec:extra_ablation} in the appendix and assess MU performance under different layer selection schemes in \textsection~\ref{subsec:ablation}. We further validate that classifier-level influence estimation generally enhances better unlearning performance. In terms of the same forget quality, it achieves 92.31\% compared to the full-model estimation result of 89.14\% in test accuracy and 8$\times$ in runtime.
\end{remark}

\begin{remark}
    In practice, $\Delta_{-\vx_i}\ell(\gD_f)$ can be heavy-tailed due to estimation noise and sample heterogeneity. Using $\psi(w)=w$ may lead to a few extreme samples dominating the weighted objective in \Cref{eqn:imu_loss}.
    We therefore introduce a smoothing transform $\psi(\cdot)$ to control the weight concentration.
    We consider several monotone choices in our experiments, including $\psi(w)=\sqrt{w}$ and $\psi(w)=\tanh(w)$, as well as a softmax-based normalization $\psi(w)=\exp(w)$. Comparison can be found in \Cref{tab:smooth} in the appendix. 
\end{remark}

\begin{remark}
Early MU methods often rely on full-parameter updates, which provide a generic solution but become impractical for large-scale models due to high compute/memory cost and unnecessary interference with retained knowledge.
Recent work, therefore, moves toward selective, parameter-efficient unlearning that updates only a small subset of parameters (\eg, SalUn~\cite{fan2023salun} and LoRA-style approaches such as LUNE~\cite{liu2025luneefficientllmunlearning}) which inject task-specific incremental parameters without modifying the base model.
Following this paradigm, \algname performs head-only unlearning, enabling efficient forgetting with minimal utility degradation.
\end{remark}

\begin{figure}[tb]
    \centering
    \includegraphics[width=0.48\textwidth]{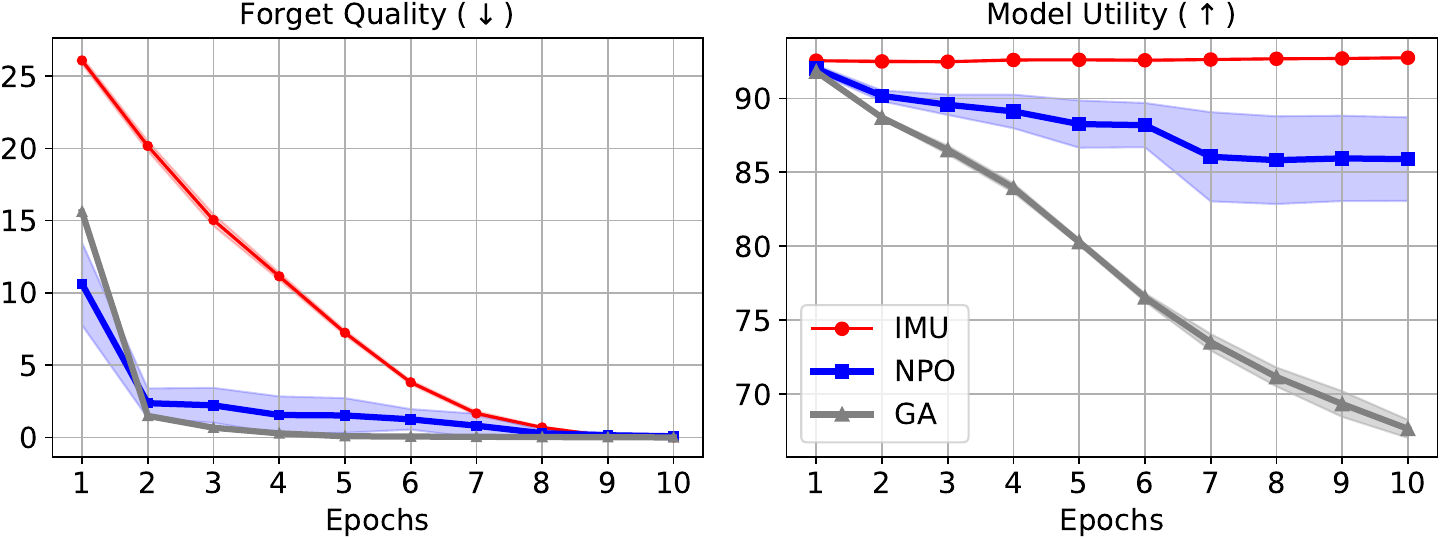}
    \caption{Divergence comparison between GA, NPO, and \algname.}
    \label{fig:npo}
\end{figure}

\subsection{Connection with NPO}
Negative Preference Optimization (NPO)~\cite{zhang2024negative} stands out in LLM unlearning, it defines
\begin{align}
    \gL_{\texttt{NPO}}(\vtheta) \coloneqq \frac{2}{\beta} \E_{\gD_f} \left[ \log \left( 1 + \left(\frac{\pi_{\vtheta}(\vy|\vx)}{\pi_{\texttt{ref}}(\vy|\vx)} \right)^\beta \right) \right],
\end{align}
where $\pi_{\vtheta}(\vy|\vx)$ denotes the output probability of the current model, and $\pi_{\texttt{ref}}(\vy|\vx)$ is the output probability from a reference model.
For image classification, the model typically outputs a softmax probability distribution over classes as $\pi_{\vtheta}(\vy|\vx) = \texttt{softmax} \left( f_{\vtheta} (\vx) \right)$, where $f_{\vtheta} (\vx)$ are the logits, and $\ell(\vx, \vy; \vtheta)=-\log \pi_{\vtheta}(\vy|\vx)$.

The gradients of GA, NPO, and \algname evaluated at $\vx_i$ are
\begin{align}
    &\vg_{\texttt{GA}} = - \nabla_{\vtheta} \ell(\vx_i, \vy_i; \vtheta) , \\
    &\vg_{\texttt{NPO}} = - w_{\texttt{NPO},i} \cdot \nabla_{\vtheta} \ell(\vx_i, \vy_i; \vtheta), \\
    &\vg_{\texttt{IMU}} = -\hat{w}_i \cdot \nabla_{\vtheta} \ell(\vx_i, \vy_i; \vtheta),
\end{align}
where $w_{\texttt{NPO},i}=2\pi_{\vtheta}(\vy|\vx)^\beta / \left(\pi_{\vtheta}(\vy|\vx)^\beta + \pi_{\texttt{ref}}(\vy|\vx)^\beta \right)$ is the adaptive smoothing weight, controlling the divergence speed.
NPO adaptively assigns weights to data points to help prevent catastrophic collapse in GA; yet, \citet{fan2024simplicity} shows that NPO exhibits reference model bias, causing an uneven allocation of unlearning power. \algname instead uses explicitly estimated influence scores to guide unlearning. In this sense, \algname can be viewed as an influence-weighted counterpart to GA, offering a more direct and interpretable mechanism for prioritizing which data points to forget.

\begin{theorem}[Divergence Analysis] \label{divergence_bound}
    Consider logistic regression with labels $y\in\{0,1\}$ and the observed-label likelihood $\pi_{\vtheta}(y\mid \vx) \triangleq \sigma \big((2y-1) z\big)$ where $z=\vtheta^\top \vx + b$ and $\sigma(u)=\frac{1}{1+\exp(-u)}$.
    Assume the model is trained on $\{(\vx_i,y_i)\}_{i=1}^{n}$ and denote $\mX=[\vx_1,\ldots,\vx_{n}]^\top\in\mathbb R^{n\times d}$, $\mK =\mX\mX^\top\in\mathbb R^{n\times n}$ be the Gram matrix.
    Then, after $t$ SGD updates with step size $\eta$, the following divergence bounds hold:
    \begin{align}
        &\|\vtheta^{t}_{\texttt{GA}}-\vtheta^{0}\|_{\mX^\top\mX}^{2} \leq \eta^2 \lambda_{\max}(\mK)^2 nt^2, \notag \\
        &\|\vtheta^{t}_{\texttt{NPO}} -\vtheta^{0}\|_{\mX^\top\mX}^{2} \le \eta^2 \lambda_{\max}(\mK)^2 nt^2 \|\mW_{\texttt{NPO}}\|^2, \notag \\
        &\|\vtheta^{t}_{\texttt{IMU}}-\vtheta^{0}\|_{\mX^\top\mX}^{2} \le \eta^2 \lambda_{\max}(\mK)^2 nt^2 \|\mW_{\texttt{IMU}}\|^2.
    \end{align}
    where $\lambda_{\max}(\mK)$ denote the largest eigenvalues of $\mK$, and $\mW_{\texttt{NPO}}\triangleq \mathrm{diag}(w_{\texttt{NPO},1},\ldots,w_{\texttt{NPO},n}), \mW_{\texttt{IMU}}\triangleq \mathrm{diag}(\hat w_{1},\ldots,\hat w_{n})$.
\end{theorem}
%
The proof of \Cref{divergence_bound} can be found in \textsection~\ref{sec:proofs}.
\Cref{divergence_bound} provides a coarse worst-case upper bound on the parameter drift under the weighted norm $\|\cdot\|_{\mX^\top\mX}$.
In particular, GA corresponds to uniform weighting, while NPO and \algname introduce per-sample weights, which can be viewed as a form of data importance weighting that reallocates unlearning strength across samples.
From this perspective, NPO can be interpreted as a confidence-based attribution (relative to the reference model), whereas \algname uses influence estimates to attribute which forgetting data are most supportive of fitting $\gD_f$ and prioritizes their removal.
Note that this result is not intended to establish a strict ordering of divergence speeds among GA, NPO, and \algname, but rather to make explicit how per-sample weighting enters the drift behavior.
We therefore complement it with empirical comparisons under tuned learning rates.

Specifically, \Cref{fig:npo} presents the class-wise forgetting on CIFAR-10.
We observe that NPO reaches low forget accuracy (around 2.37\%) quickly and then stabilizes, but it is accompanied by a continuous decline in model utility, from 92.01\% to 85.89\%.
In contrast, \algname exhibits a more gradual decrease in forget accuracy while consistently maintaining a higher test accuracy around 92.60\%.
Further comparisons on language tasks are reported in \textsection~\ref{sec:exp} and \textsection~\ref{sec: extra_res}.

\begin{table*}[tb]
\caption{Quantitative results on CIFAR-10 and CIFAR-100. Results are averaged over all 10 classes for class-wise unlearning on CIFAR-10, across 20 superclasses, and across 5 subclasses in one superclass for subclass-wise unlearning on CIFAR-100.}
  \centering
  \label{tab:class-wise}
  \begin{adjustbox}{max width=0.98\textwidth}
  \begin{tabular}{llcccccccc}
    \toprule
    Setting &Method  & $\gD_r$ & $\gD_f$ 
                 & $\texttt{Acc}_{\gD_f} (\downarrow)$ 
                 & $\texttt{ACC}_{\gD_r} (\uparrow)$ 
                 & $\texttt{ACC}_{\gD_t} (\uparrow)$ 
                 & MIA $(\uparrow)$
                 & $W_\texttt{dist} (\downarrow)$ 
                 & Run time (s) $(\downarrow)$  \\
    \midrule
    \multirow{8}{*}{CIFAR-10}
    &Original     & \cmark     & \cmark   & 
    99.71\scriptsize{$\pm$0.00} & 
    \phantom{0}99.45\scriptsize{$\pm$0.00\phantom{0}} & \phantom{0}94.43\scriptsize{$\pm$0.00\phantom{0}} & \phantom{0}0.01\scriptsize{$\pm$0.00} & –  & – \\
    &Retrain      & \cmark     & \xmark   & \phantom{0}0.00\scriptsize{$\pm$0.00}  & \phantom{0}99.98\scriptsize{$\pm$0.00\phantom{0}} & \phantom{0}94.41\scriptsize{$\pm$0.35\phantom{0}} & \phantom{0}1.00\scriptsize{$\pm$0.00} & \phantom{0}0.00\scriptsize{$\pm$0.00} & – \\
    \cmidrule{2-10}
    &IU$^\ast$       & \cmark     & \cmark   & \phantom{0}7.56\scriptsize{$\pm$1.43}  & \phantom{0}96.04\scriptsize{$\pm$0.95\phantom{0}} & \phantom{0}90.24\scriptsize{$\pm$0.91\phantom{0}} & \phantom{0}0.96\scriptsize{$\pm$0.03} & \phantom{0}0.15\scriptsize{$\pm$0.07} & \phantom{0}55\scriptsize{$\pm$1} \\
    &GA        & \xmark     & \cmark   & \phantom{0}0.48\scriptsize{$\pm$0.10}  & \phantom{0}75.89\scriptsize{$\pm$6.28\phantom{0}} & \phantom{0}72.43\scriptsize{$\pm$5.83\phantom{0}} & \phantom{0}0.99\scriptsize{$\pm$0.00} & \phantom{0}3.59\scriptsize{$\pm$0.39} & \phantom{0}29\scriptsize{$\pm$0} \\
    &RL           & \xmark     & \cmark   & \phantom{0}0.71\scriptsize{$\pm$0.42}  & \phantom{0}79.01\scriptsize{$\pm$13.25} & \phantom{0}73.33\scriptsize{$\pm$12.25} & \phantom{0}0.99\scriptsize{$\pm$0.00} & \phantom{0}4.04\scriptsize{$\pm$0.61} & \phantom{0}25\scriptsize{$\pm$5} \\
    &SCAR$^\ast$ & \xmark     & \cmark   & \phantom{0}0.50\scriptsize{$\pm$1.03}  & \phantom{0}85.78\scriptsize{$\pm$4.15\phantom{0}} & \phantom{0}80.99\scriptsize{$\pm$4.07\phantom{0}} & \phantom{0}1.00\scriptsize{$\pm$0.00} & \phantom{0}4.80\scriptsize{$\pm$0.62} & 325\scriptsize{$\pm$11} \\
    
    &NPO        & \xmark     & \cmark   & \phantom{0}0.20\scriptsize{$\pm$0.28}  & \phantom{0}91.63\scriptsize{$\pm$7.23} & \phantom{0}85.84\scriptsize{$\pm$6.66\phantom{0}} & \phantom{0}1.00\scriptsize{$\pm$0.00} & \phantom{0}0.24\scriptsize{$\pm$0.03} & \phantom{0}39\scriptsize{$\pm$2} \\

     & \cellcolor{gray!20}\algname (ours)  & \cellcolor{gray!20}\xmark    & \cellcolor{gray!20}\cmark   &\cellcolor{gray!20} \textbf{\phantom{0}0.18\scriptsize{$\pm$0.27}} &\cellcolor{gray!20}  \textbf{\phantom{0}98.93\scriptsize{$\pm$0.30\phantom{0}}} &\cellcolor{gray!20} \textbf{\phantom{0}93.38\scriptsize{$\pm$1.06\phantom{0}}} &\cellcolor{gray!20} \textbf{\phantom{0}1.00\scriptsize{$\pm$0.00}}   &\cellcolor{gray!20} \textbf{\phantom{0}0.07\scriptsize{$\pm$0.02}} &\cellcolor{gray!20} \phantom{0}58\scriptsize{$\pm$0}\\
    \midrule
    \multirow{9}{*}{CIFAR-100}
    
    &Original     & \cmark     & \cmark   & 99.89\scriptsize{$\pm$0.19} & \phantom{0}99.99\scriptsize{$\pm$0.02} & \phantom{0}76.63\scriptsize{$\pm$5.47} & \phantom{0}0.01\scriptsize{$\pm$0.00} & – & – \\
    &Retrain      & \cmark     & \xmark   & \phantom{0}0.00\scriptsize{$\pm$0.00}  & \phantom{0}99.53\scriptsize{$\pm$2.79} & \phantom{0}74.05\scriptsize{$\pm$8.95} & \phantom{0}1.00\scriptsize{$\pm$0.00} & \phantom{0}0.00\scriptsize{$\pm$0.00} & – \\
    \cmidrule{2-10}
    &IU$^\ast$  & \cmark     & \cmark   & \phantom{0}0.00\scriptsize{$\pm$0.00}  & \phantom{0}69.29\scriptsize{$\pm$4.00} & \phantom{0}61.85\scriptsize{$\pm$5.18} & \phantom{0}1.00\scriptsize{$\pm$0.00} & \phantom{0}4.82\scriptsize{$\pm$2.00} & 19\scriptsize{$\pm$2} \\
    &GA   & \xmark     & \cmark   & \phantom{0}0.00\scriptsize{$\pm$0.00}  & \phantom{0}68.29\scriptsize{$\pm$5.27} & \phantom{0}40.89\scriptsize{$\pm$8.61} & \phantom{0}1.00\scriptsize{$\pm$0.00} & \phantom{0}5.22\scriptsize{$\pm$0.14} & \phantom{0}9\scriptsize{$\pm$0} \\
    &RL  & \xmark     & \cmark   & \phantom{0}0.00\scriptsize{$\pm$0.00}  & \phantom{0}73.87\scriptsize{$\pm$8.81} & \phantom{0}42.80\scriptsize{$\pm$7.68} & \phantom{0}1.00\scriptsize{$\pm$0.00} & \phantom{0}7.73\scriptsize{$\pm$0.81} & 22\scriptsize{$\pm$3} \\
    &SSD  & \cmark     & \cmark   & \phantom{0}0.00\scriptsize{$\pm$0.00}  & \phantom{0}77.37\scriptsize{$\pm$9.93} & \phantom{0}46.85\scriptsize{$\pm$2.90} & \phantom{0}1.00\scriptsize{$\pm$0.00} & \phantom{0}3.15\scriptsize{$\pm$1.27} & 10\scriptsize{$\pm$0} \\
    &SCAR$^\ast$   & \xmark     & \cmark   & \phantom{0}0.00\scriptsize{$\pm$0.00}  & \phantom{0}77.79\scriptsize{$\pm$1.28} & \phantom{0}60.00\scriptsize{$\pm$1.48} & \phantom{0}1.00\scriptsize{$\pm$0.00} & \phantom{0}7.13\scriptsize{$\pm$1.51} & 68\scriptsize{$\pm$0} \\
    &NPO  & \xmark     & \cmark   & \phantom{0}0.00\scriptsize{$\pm$0.00}  & \phantom{0}90.47\scriptsize{$\pm$3.15} & \phantom{0}55.80\scriptsize{$\pm$7.55} & \phantom{0}1.00\scriptsize{$\pm$0.00} & \phantom{0}5.35\scriptsize{$\pm$1.10} & 25\scriptsize{$\pm$3} \\
    &\cellcolor{gray!20} \algname (ours)    &\cellcolor{gray!20}  \xmark     &\cellcolor{gray!20}  \cmark   &\cellcolor{gray!20}  \textbf{\phantom{0}0.00\scriptsize{$\pm$0.00}}  & \cellcolor{gray!20} \textbf{\phantom{0}98.06\scriptsize{$\pm$1.35}} &\cellcolor{gray!20}  \textbf{\phantom{0}63.75\scriptsize{$\pm$7.85}} &\cellcolor{gray!20}  \textbf{\phantom{0}1.00\scriptsize{$\pm$0.00}} &\cellcolor{gray!20}  \textbf{\phantom{0}2.09\scriptsize{$\pm$0.37}} &\cellcolor{gray!20}  14\scriptsize{$\pm$0} \\
    \midrule
    \multirow{9}{*}{CIFAR-100}
    
    &Original     & \cmark     & \cmark   & 100.00\scriptsize{$\pm$0.00} & 100.00\scriptsize{$\pm$0.00\phantom{0}} & \phantom{0}77.80\scriptsize{$\pm$2.98} & \phantom{0}0.00\scriptsize{$\pm$0.00} & – & – \\
    &Retrain      & \cmark     & \xmark   & \phantom{0}0.00\scriptsize{$\pm$0.00\phantom{0}}  & \phantom{0}99.33\scriptsize{$\pm$1.16\phantom{0}} & \phantom{0}76.49\scriptsize{$\pm$2.53} & \phantom{0}1.00\scriptsize{$\pm$0.00} & \phantom{0}0.00\scriptsize{$\pm$0.00} & – \\
    \cmidrule{2-10}
    &IU$^\ast$   & \cmark     & \cmark   & \phantom{0}0.00\scriptsize{$\pm$0.00\phantom{0}}  & \phantom{0}60.58\scriptsize{$\pm$4.89\phantom{0}} & \phantom{0}54.50\scriptsize{$\pm$5.31} & \phantom{0}1.00\scriptsize{$\pm$0.00} & \phantom{0}5.80\scriptsize{$\pm$0.08} & 18\scriptsize{$\pm$0} \\
    &GA    & \xmark     & \cmark   & \phantom{0}0.00\scriptsize{$\pm$0.00\phantom{0}}  & \phantom{0}75.56\scriptsize{$\pm$9.46\phantom{0}} & \phantom{0}49.50\scriptsize{$\pm$6.16} & \phantom{0}1.00\scriptsize{$\pm$0.00} & \phantom{0}7.54\scriptsize{$\pm$1.29} & \phantom{0}9\scriptsize{$\pm$0} \\
    &RL   & \xmark     & \cmark   & \phantom{0}0.00\scriptsize{$\pm$0.00\phantom{0}}  & \phantom{0}76.49\scriptsize{$\pm$11.51} & \phantom{0}47.69\scriptsize{$\pm$4.54} & \phantom{0}1.00\scriptsize{$\pm$0.00} & \phantom{0}7.57\scriptsize{$\pm$0.07} & 19\scriptsize{$\pm$2} \\
    &SSD    & \cmark     & \cmark   & \phantom{0}0.00\scriptsize{$\pm$0.00\phantom{0}}  & \phantom{0}83.07\scriptsize{$\pm$1.13\phantom{0}} & \phantom{0}52.20\scriptsize{$\pm$1.32} & \phantom{0}1.00\scriptsize{$\pm$0.00} & \phantom{0}4.81\scriptsize{$\pm$0.50} & 10\scriptsize{$\pm$0} \\
    &SCAR$^\ast$   & \xmark     & \cmark   & \phantom{0}0.00\scriptsize{$\pm$0.00\phantom{0}}  & \phantom{0}78.72\scriptsize{$\pm$4.80\phantom{0}} & \phantom{0}56.55\scriptsize{$\pm$5.31} & \phantom{0}1.00\scriptsize{$\pm$0.00} & \phantom{0}5.88\scriptsize{$\pm$0.25} & 67\scriptsize{$\pm$0} \\
    &NPO  & \xmark     & \cmark   & \phantom{0}0.00\scriptsize{$\pm$0.00\phantom{0}}  & \phantom{0}90.22\scriptsize{$\pm$4.57\phantom{0}} & \phantom{0}58.90\scriptsize{$\pm$6.34} & \phantom{0}1.00\scriptsize{$\pm$0.00} & \phantom{0}5.61\scriptsize{$\pm$0.25} & 26\scriptsize{$\pm$2} \\
    &\cellcolor{gray!20} \algname (ours)    & \cellcolor{gray!20} \xmark     &\cellcolor{gray!20}  \cmark   &\cellcolor{gray!20}  \textbf{\phantom{0}0.00\scriptsize{$\pm$0.00\phantom{0}}}  &\cellcolor{gray!20}  \textbf{\phantom{0}98.61\scriptsize{$\pm$0.85\phantom{0}}} &\cellcolor{gray!20}  \textbf{\phantom{0}67.60\scriptsize{$\pm$3.44}} & \cellcolor{gray!20} \textbf{\phantom{0}1.00\scriptsize{$\pm$0.00}} &\cellcolor{gray!20}  \textbf{\phantom{0}3.69\scriptsize{$\pm$0.16}} &\cellcolor{gray!20}  14\scriptsize{$\pm$0} \\
    \bottomrule
  \end{tabular}
  \end{adjustbox}

\textit{Note:} IU$^\ast$ is an improved version presented in the paper~\cite{fan2023salun}, SCAR$^\ast$ denotes the version using $\gD_f$ only.
\end{table*}

\section{Experiment}
\label{sec:exp}



\subsection{Setup}

\noindent
\textbf{Dataset \& models}.
We evaluate on image classification using CIFAR-10 and CIFAR-100~\cite{krizhevsky2009learning} with ResNet-18~\cite{he2016deep}.
Person re-identification, which matches a person's identity across different cameras or locations in a video or image sequence, is also included.
ResNet-50 with a fully-connected layer is trained on the Market-1501 dataset~\cite{zheng2015scalable}, which contains labeled person IDs captured under varying camera views.
We also assess MU in a sequence modeling setting using a GPT-2 model on synthetic data~\cite{fan2024simplicity}.
For LLM unlearning, we employ the Llama-3.2-3B-Instruct model on TOFU~\cite{openunlearning2025}.

\noindent
\textbf{Scenarios}.
We benchmark across five different unlearning settings:
(\romannumeral1) Class-wise unlearning: forget a full category from CIFAR-10.
(\romannumeral2) Subclass-wise unlearning: forget a subclass within a superclass (\eg, boy from people) in CIFAR-100.
(\romannumeral3) Sample-wise unlearning (see appendix): randomly select samples from $\gD_{\texttt{train}}$ as the forget set.
(\romannumeral4) Person re-id unlearning: forget a specific identity from the 751 IDs in the Market-1501 train set, and evaluate model generalization on the query and gallery sets.
(\romannumeral5) Distributional unlearning in sequence modeling (see appendix): forget two Markov sub-distributions in a synthetic task. 
(\romannumeral6) LLM unlearning (see appendix): perform forget05 task on TOFU.

\noindent
\textbf{Baselines}. (1) Retrain, (2) Gradient Ascent (GA)~\cite{thudi2022unrolling},  (3) Random Label (RL)~\cite{golatkar2020eternal}, (4) Influence Unlearning (IU$^\ast$)~\cite{fan2023salun}, 
(5) SCAR$^\ast$~\cite{bonato2024retain},
(6) SSD~\cite{foster2024fast}, (7) NPO~\cite{zhang2024negative}, (8) SimNPO~\cite{fan2024simplicity}.

\noindent
\textbf{Metrics}.
(1) Accuracy on $\gD_f$, $\gD_r$, and $\gD_t$, denoted as $\texttt{Acc}_{\gD_f}$, $\texttt{Acc}_{\gD_r}$, and $\texttt{Acc}_{\gD_t}$, respectively.
(2) Membership inference attacks (\textit{MIA})~\cite{fan2023salun} aim to infer information about the training data. 
(3) \textit{$W_\texttt{dist}$}~\cite{tarun2023deep}, \ie, Wasserstein-1 distance, by measuring the similarity between the output distributions of the unlearned model and the retrained model on $\gD_r$.
(4) \textit{mAP} (mean average precision) represents the average of the area under the precision-recall curve for each query, reflecting the overall retrieval quality.

\subsection{Main results}

\noindent
\textbf{Results on class-wise unlearning.}
We first evaluate on CIFAR-10, trying to forget an entire class category.
As shown in \Cref{tab:class-wise}, our proposed method \algname achieves superior unlearning performance.

IU$^\ast$ fails to effectively preserve the model utility, showing significant drops of $\sim$15\% in accuracy in both $\gD_r$ and the test set $\gD_t$. GA and RL perform even worse in this regard.
The SOTA retain-data-free MU method SCAR$^\ast$ demonstrates strong forgetting ability; however, it comes at the cost of notable degradation in model utility, with accuracies on $\gD_r$ and $\gD_t$ reduced by approximately 14\%.
We further employ the potent LLM unlearning method, NPO, for image classification. It significantly outperforms baselines in both the forget quality and model utility preservation.

Similarly, our proposed method \algname, not only achieves strong forget capabilities, \ie, 0.02\% on $\gD_f$, but also best preserves the model utility, with only about a 2\% drop on both $\gD_r$ and $\gD_t$, indicating that \algname strikes a desirable balance between effective forgetting and minimal impact on model utility.
In addition, \algname demonstrates favorable efficiency, is over 3$\times$ faster than SCAR$^\ast$.

\noindent
\textbf{Results on subclass-wise unlearning.}
We further evaluate MU methods on CIFAR-100, trying to forget a single subclass from a semantic superclass.
Specifically, (1) for subclass-wise unlearning across all super-classes, we randomly select a subclass from each of the 20 super-classes in CIFAR-100 to be forgotten, and report the average performance;
(2) for subclass-wise unlearning within a single super-class, we focus on the super-class `people’, where each time, one of the five subclasses is selected to be forgotten while the remaining four are retained. 
These settings are more challenging than the standard class-wise unlearning, as subclasses within a superclass share semantically and visually similar features, making the forgetting more fine-grained and less separable.

Despite the increased difficulty, \algname consistently demonstrates better performance than other methods. In \Cref{tab:class-wise}, all MU methods completely remove the knowledge about $\gD_f$, while \algname best preserving the performance over $\gD_r$, having an accuracy around $98\%$ and achieves a high test accuracy compared to other baselines.

\begin{table}[tb]
\aboverulesep = 0pt
\abovetopsep = 0pt
\belowrulesep = 0pt 
\caption{Unlearning on person re-identification.}
\centering 
\label{tab:pid}
\begin{adjustbox}{max width=0.48\textwidth}
\begin{tabular}{lcccc}
    \toprule
    Method         & mAP ($\uparrow$)   & Top-1 ($\uparrow$) & Top-5 ($\uparrow$) & Run time ($\downarrow$)\\
    \midrule
    Original       & 68.50 &85.45 & 94.27 & --       \\
    \cmidrule{1-5}
    IU$^\ast$ & 44.79 & 69.69 & 84.59 & 147      \\
    GA & \phantom{0}4.12 & 18.59 & 32.66 & \phantom{0}43      \\
    RL & \phantom{0}0.83 & \phantom{0}0.02 & \phantom{0}0.06 &104      \\
    SSD & 51.79 & 73.57 & 87.35 & 137      \\
    SCAR$^\ast$ & 50.97 & 73.87 & 88.54 & 135\\
    NPO & 41.40 & 63.13 & 81.03 & \phantom{0}55      \\
    \rowcolor{gray!20} \algname (Ours)  & \textbf{55.85} & \textbf{76.07} & \textbf{88.75} & \phantom{0}43 \\
    \bottomrule
\end{tabular}
\end{adjustbox}

\end{table}
%

\begin{figure}[tb]
    \centering
    \includegraphics[width=0.40\textwidth]{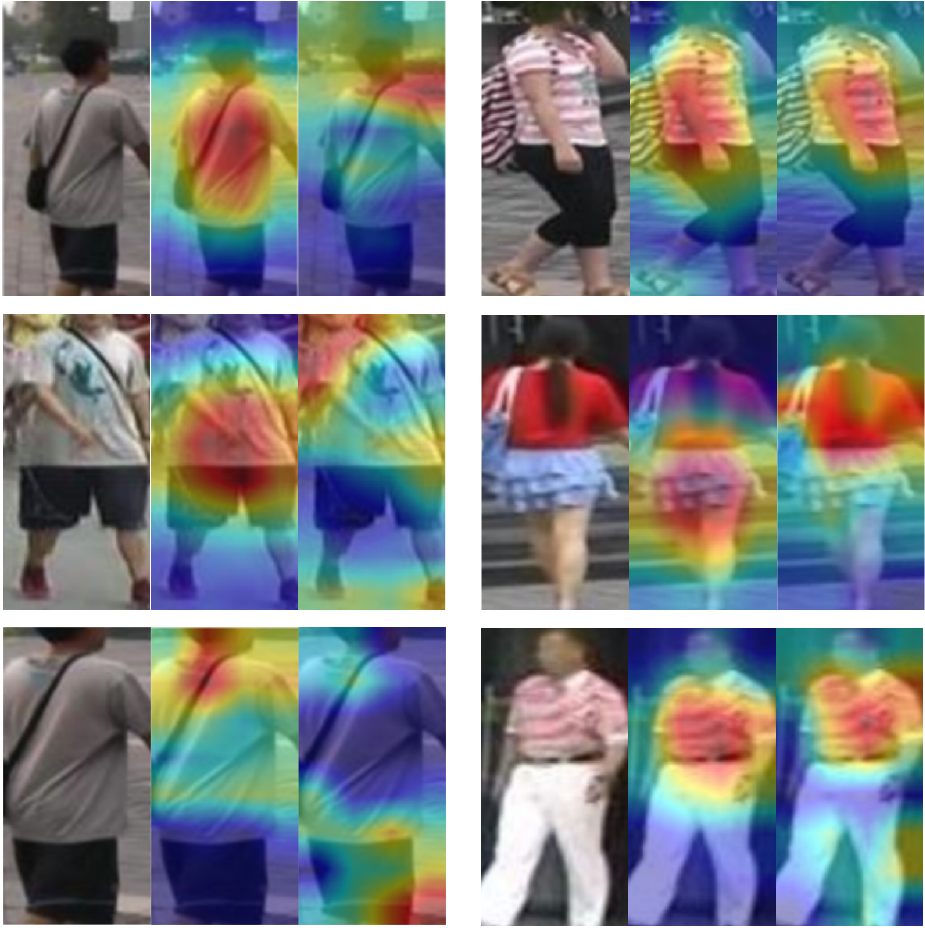}
    \caption{Visualizations of regions where models focus on for $\gD_f$ and $\gD_r$, respectively. 
    For each triplet, from left to right are the original image, the activation map generated by the original model, and \algname scrubbed model, respectively.}
    \label{fig:pid}
\end{figure}

\noindent
\textbf{Results on person re-identification.}
Pedestrians' body shapes, clothing, facial features, and other biometric characteristics are often sensitive and uniquely identifiable, making them important targets for protection in MU scenarios. To this end, we also evaluate MU methods on the person re-identification task. This task involves matching images of the same individual captured under varying camera viewpoints or multiple cameras. Here, we aim to forget all samples associated with a particular identity (\eg, $\texttt{pid}=1$).
This task poses unique challenges to MU methods, as forgetting a specific individual requires precise removal of identity-related features while preserving general person-level recognition performance.
We adopt GradCAM~\cite{selvaraju2017grad} to visualize regions where models focus on \without and \with \algname.
As shown in \Cref{fig:pid}, when evaluated on $\gD_r$, \algname preserves focus on discriminative yet identity-agnostic regions. In contrast, when evaluated on $\gD_f$, our scrubbed model significantly shifts attention away from identity-revealing regions (\eg, clothing logos, face, and hairstyle), indicating successful removal of sensitive cues.
In \Cref{tab:pid}, we further evaluate model utility by measuring mAP and Top-$k$ accuracy on a query-gallery split consisting of individuals entirely unseen during training.
In general, \algname achieves a good trade-off between forget quality and model utility.

\noindent
\textbf{Computational complexity analysis.}
Also note that, for a linear softmax classifier trained with cross-entropy, the Hessian \wrt the head parameters admits a closed form as
\begin{align}
  \mH_{\vtheta_c^\ast} = \frac{1}{|\gD_f|} \sum_{i=1}^{|\gD_f|} \mJ_i^\top \left[ \operatorname{diag}(\vp_i) - \vp_i \vp_i^\top \right] \mJ_i,
\end{align}
where $\vp_i = \mathrm{softmax}(f_{\vtheta_c^\ast}(\vx_i))$ denotes the predicted probability vector
and $\mJ_i = \frac{\partial f_{\vtheta_c^\ast}(\vx_i)}{\partial \vtheta_c}$ is the Jacobian of the last classifier head.
In practice, to mitigate runtime and memory demands, we adopt a Fisher diagonal approximation in implementation and compute it only for the head (size $d$) over $N$ forgetting data every epoch, so the extra overhead of \algname over GA is $\mathcal{O}(NdC)$ where $C$ is the number of classes. 
Note that all run times reported in our tables include the full pipeline end-to-end, encompassing influence score computation, sample selection and reweighting, as well as the gradient ascent unlearning updates.
In experiments, \algname requires on average around twice the run time of GA under the same number of training epochs, indicating that the additional computational cost introduced by the influence estimation stage remains within an acceptable range.


\subsection{Ablation study}
\label{subsec:ablation}
We finally conduct ablation studies in the class-wise unlearning scenario on CIFAR-10 to analyze the effects of key design choices in our method, including (1) the ratio $r$ of top-ranked forgetting data points selected based on the influence values, (2) the update frequency $\nu$ of influence scores during the unlearning process, (3) the smoothing strategy applied to the influence $\gI^c(\vx_i, \gD_f)$, (4) the dataset used for influence estimation,
and (5) the unfreeze strategy of model layers, as well as the choice of layer for computing influence values. Ablations (3)-(5) can be found in \textsection~\ref{sec: extra_res} in the appendix.

%
\Cref{tab:ratio} presents the effect of varying the top-$r$ fraction of the forgetting data $\gD_f$ for unlearning. As $r$ decreases, the model retains strong forget quality as evidenced by near-zero accuracy on $\gD_f$, while achieving slightly improved accuracy on $\gD_r$ and $\gD_t$.
Notably, using only 5\% of the most influential data points is sufficient to induce forgetting, while also enhancing generalization and utility, highlighting the efficiency of targeting high-impact data points.

\Cref{tab:frequency} investigates the impact of varying the frequency $\nu$ of influence score updates.
All variants achieve strong forget quality, with minor differences in model utility. Infrequent updates are sufficient for maintaining effective unlearning, indicating that the approximation error introduced by using fixed influence weights remains negligible even as the feature space evolves. To maintain fast execution time, we set frequency $\nu=0$ in main experiments.

Additional results provided in the appendix show that different smoothing strategies for $\gI^c(\vx_i, \gD_f)$ (e.g., \texttt{sqrt}, \texttt{softmax}, \texttt{tanh}) yield similar outcomes, with \texttt{sqrt} performing slightly better overall. Estimating the influence value using only $\gD_f$ is sufficient compared to using $\gD_{train}$. Moreover, updating only the last classifier layer achieves the best balance between forgetting and preservation, while unfreezing earlier layers or computing influence across all layers significantly degrades performance and increases runtime.

\begin{table}[tb]
\caption{Effect of varying the ratio ($r$) of top-ranked $\gD_f$ (selected based on $\gI^c(\vx, \gD_f)$) on unlearning performance.}
\centering
\label{tab:ratio}
\begin{adjustbox}{max width=0.48\textwidth}
\begin{sc}
\begin{tabular}{lccccc }
    \toprule
    $r$ &$\texttt{Acc}_{\gD_f} (\downarrow)$ &$\texttt{Acc}_{\gD_r} (\uparrow)$ &$\texttt{Acc}_{\gD_t} (\uparrow)$ & MIA $(\uparrow)$ &Run time $(\downarrow)$\\
    \midrule
    1.00  & 0.02\scriptsize{$\pm$0.03}       & 97.93\scriptsize{$\pm$0.84}      & 91.22\scriptsize{$\pm$0.77}    & 1.00\scriptsize{$\pm$0.00} &56\scriptsize{$\pm$0} \\
    0.80  & 0.18\scriptsize{$\pm$0.32}       & 97.89\scriptsize{$\pm$0.80}      & 91.19\scriptsize{$\pm$0.58}    & 1.00\scriptsize{$\pm$0.00} &54\scriptsize{$\pm$0} \\
    0.60   & 0.04\scriptsize{$\pm$0.08}       & 97.94\scriptsize{$\pm$0.86}      & 91.32\scriptsize{$\pm$0.62}    & 1.00\scriptsize{$\pm$0.00} &53\scriptsize{$\pm$0} \\
    0.40  & 0.03\scriptsize{$\pm$0.05}       & 98.00\scriptsize{$\pm$0.71}      & 91.14\scriptsize{$\pm$0.83}    & 1.00\scriptsize{$\pm$0.00} &51\scriptsize{$\pm$0} \\
    
    0.20  & 0.00\scriptsize{$\pm$0.00}       & 98.62\scriptsize{$\pm$0.70}      & 91.98\scriptsize{$\pm$0.53}    & 1.00\scriptsize{$\pm$0.00} &50\scriptsize{$\pm$0} \\
    0.05  & 0.00\scriptsize{$\pm$0.00}       & 99.79\scriptsize{$\pm$0.19}      & 93.78\scriptsize{$\pm$0.24}    & 1.00\scriptsize{$\pm$0.00} &48\scriptsize{$\pm$0} \\
    \bottomrule
\end{tabular}
\end{sc}
\end{adjustbox}
\end{table}
\begin{table}[tb]
\caption{Effect of frequency ($\nu$) of influence value updates on unlearning performance.
$\nu=0$ means only update at the first epoch; $\nu=1$ and $\nu=2$ mean update every epoch and every two epochs, respectively.}
\centering
\label{tab:frequency}
\begin{adjustbox}{max width=0.48\textwidth}
\begin{sc}
\begin{tabular}{lccccc}
\toprule
$\nu$ &$\texttt{Acc}_{\gD_f} (\downarrow)$ &$\texttt{Acc}_{\gD_r} (\uparrow)$ &$\texttt{Acc}_{\gD_t} (\uparrow)$ & MIA $(\uparrow)$  & Run time $(\downarrow)$ \\
\midrule
$0$  & 0.02\scriptsize{$\pm$0.02}       & 97.75\scriptsize{$\pm$0.70}      & 91.86\scriptsize{$\pm$0.53}    & 1.00\scriptsize{$\pm$0.00} &\phantom{0}56\scriptsize{$\pm$0} \\
$1$  & 0.00\scriptsize{$\pm$0.00}       & 97.65\scriptsize{$\pm$0.73}      & 91.71\scriptsize{$\pm$0.54}    & 1.00\scriptsize{$\pm$0.00} &231\scriptsize{$\pm$1} \\ 
$2$  & 0.00\scriptsize{$\pm$0.00}       & 97.72\scriptsize{$\pm$0.73}      & 91.82\scriptsize{$\pm$0.58}    & 1.00\scriptsize{$\pm$0.00} &175\scriptsize{$\pm$3} \\ 
\bottomrule
\end{tabular}
\end{sc}
\end{adjustbox}
\end{table}

\section{Conclusion and Limitations}

In this paper, we introduce \algname, a novel retain-data-free MU method that dynamically allocates optimization priority to each data point according to its estimated influence scores.
By leveraging a computationally efficient classifier-level influence approximation, \algname bypasses the prohibitive overhead of full-model Hessian inversion while maintaining high fidelity to the true influence signals.
Extensive experiments across diverse unlearning scenarios in both vision and language tasks demonstrate that \algname effectively strikes the balance between forget quality and model utility, consistently outperforming existing baselines.

\noindent
\textbf{Limitation.}
While \algname requires computing influence scores for each forgetting data point, the cost scales with the size of $\gD_f$. A natural mitigation is to estimate influence at the mini-batch level; however, this may introduce additional approximation error and is left for future investigation.

While \algname's default head-only update prioritizes efficiency, it may leave frozen backbone representations partially vulnerable to sophisticated adversarial probing.
However, we emphasize that the influence-guided reweighting paradigm at the core of \algname is inherently strategy-agnostic.
It can be seamlessly integrated with more intensive update schemes, such as full-parameter fine-tuning or adaptive layer unfreezing (see \textsection~\ref{subsec:extra_ablation} in the appendix), depending on the specific requirements for unlearning depth and defensive robustness.
Exploring these hybrid extensions under adversarial settings constitutes a promising direction for future research.


\bibliographystyle{ACM-Reference-Format}
\bibliography{acmref}

%
\appendix
\newpage

\setcounter{page}{1}
\onecolumn
{
\centering
\Large
\textbf{IMU: Influence-guided Machine Unlearning}\\
\vspace{0.5em}Supplementary Material \\
\vspace{1.0em}
}

\phantomsection
\addcontentsline{toc}{part}{Appendix} 

\section{Proofs}
\subsection{Classifier-Level Influence Approximation}
\label{classifier-level proof}

\textbf{Theorem 3.4}(Classifier-Level Loss-Influence Approximation). 
Let $\gL_f(\vtheta)\triangleq \ell(f_\vtheta(\gD_f))$, and let the Hessian of $\gL_f$ at $\vtheta^*$ be
$$
\mH^f_{\vtheta^*}
=
\begin{bmatrix}
\mH_{ee} & \mH_{ec} \\
\mH_{ce} & \mH_{cc}
\end{bmatrix}.
$$
Define
\begin{align}
\Delta^{\mathrm{full}}_{-\vx}\ell(\gD_f)
&\triangleq
\frac{1}{N}
\nabla_{\vtheta}\gL_f(\vtheta^*)^\top
(\mH^f_{\vtheta^*})^{-1}
\nabla_{\vtheta}\ell(f_{\vtheta^*}(\vx)), \\
\Delta^{c}_{-\vx}\ell(\gD_f)
&\triangleq
\frac{1}{N}
\nabla_{\vtheta_c}\gL_f(\vtheta^*)^\top
\mH_{cc}^{-1}
\nabla_{\vtheta_c}\ell(f_{\vtheta^*}(\vx)).
\end{align}
Under \Cref{assumption}, we have
\begin{align}
\Big|
\Delta^{\mathrm{full}}_{-\vx}\ell(\gD_f)
-
\Delta^{c}_{-\vx}\ell(\gD_f)
\Big|
=
\mathcal{O}\!\Bigg(
\frac{1}{N}
\Big(
&\|\mH_{ce}\mH_{ee}^{-1}\mH_{ec}\|
+
\|\nabla_{\vtheta_e}\gL_f(\vtheta^*)\|\,\|\mH_{ee}^{-1}\mH_{ec}\| \notag\\
&+
\|\mH_{ce}\mH_{ee}^{-1}\|\,\|\nabla_{\vtheta_e}\ell(f_{\vtheta^*}(\vx))\|
+
\|\nabla_{\vtheta_e}\gL_f(\vtheta^*)\|\,\|\nabla_{\vtheta_e}\ell(f_{\vtheta^*}(\vx))\|
\Big)
\Bigg).
\end{align}

\begin{proof}
Let
$$
\mathbf u
=
\begin{bmatrix}
\mathbf u_e\\
\mathbf u_c
\end{bmatrix}
=
\nabla_{\vtheta}\gL_f(\vtheta^*),
\qquad
\mathbf v
=
\begin{bmatrix}
\mathbf v_e\\
\mathbf v_c
\end{bmatrix}
=
\nabla_{\vtheta}\ell(f_{\vtheta^*}(\vx)),
$$
where $\mathbf u_e=\nabla_{\vtheta_e}\gL_f(\vtheta^*)$, $\mathbf u_c=\nabla_{\vtheta_c}\gL_f(\vtheta^*)$, $\mathbf v_e=\nabla_{\vtheta_e}\ell(f_{\vtheta^*}(\vx))$, and $\mathbf v_c=\nabla_{\vtheta_c}\ell(f_{\vtheta^*}(\vx))$.
Define
$$
\mE \triangleq \mH_{ce}\mH_{ee}^{-1}\mH_{ec},
\qquad
\mS \triangleq \mH_{cc}-\mE.
$$
By the block inverse formula,
$$
(\mH^f_{\vtheta^*})^{-1}
=
\begin{bmatrix}
\mH_{ee}^{-1}+\mH_{ee}^{-1}\mH_{ec}\mS^{-1}\mH_{ce}\mH_{ee}^{-1}
&
-\mH_{ee}^{-1}\mH_{ec}\mS^{-1}
\\[3pt]
-\mS^{-1}\mH_{ce}\mH_{ee}^{-1}
&
\mS^{-1}
\end{bmatrix}.
$$
Under \Cref{assumption}, $\|\mH_{cc}^{-1}\mE\|\ll 1$, and thus the Neumann expansion gives
$$
\mS^{-1}
=
(\mH_{cc}-\mE)^{-1}
=
\mH_{cc}^{-1}
+
\mH_{cc}^{-1}\mE\mH_{cc}^{-1}
+
\mathcal{O}\!\left(\|\mH_{cc}^{-1}\mE\|^2\right),
$$
which implies
$$
\mS^{-1}
=
\mH_{cc}^{-1}
+
\mathcal{O}(\|\mE\|).
$$

Now expand the bilinear form:
\begin{align}
\mathbf u^\top (\mH^f_{\vtheta^*})^{-1}\mathbf v
=
&\mathbf u_c^\top \mS^{-1}\mathbf v_c
-\mathbf u_e^\top \mH_{ee}^{-1}\mH_{ec}\mS^{-1}\mathbf v_c \notag\\
&-\mathbf u_c^\top \mS^{-1}\mH_{ce}\mH_{ee}^{-1}\mathbf v_e
+\mathbf u_e^\top
\left(
\mH_{ee}^{-1}
+\mH_{ee}^{-1}\mH_{ec}\mS^{-1}\mH_{ce}\mH_{ee}^{-1}
\right)\mathbf v_e.
\end{align}
Therefore,
\begin{align}
\mathbf u^\top (\mH^f_{\vtheta^*})^{-1}\mathbf v
-
\mathbf u_c^\top \mH_{cc}^{-1}\mathbf v_c
=
&\mathbf u_c^\top(\mS^{-1}-\mH_{cc}^{-1})\mathbf v_c \notag\\
&-\mathbf u_e^\top \mH_{ee}^{-1}\mH_{ec}\mS^{-1}\mathbf v_c \notag\\
&-\mathbf u_c^\top \mS^{-1}\mH_{ce}\mH_{ee}^{-1}\mathbf v_e \notag\\
&+\mathbf u_e^\top
\left(
\mH_{ee}^{-1}
+\mH_{ee}^{-1}\mH_{ec}\mS^{-1}\mH_{ce}\mH_{ee}^{-1}
\right)\mathbf v_e.
\end{align}

For the first term, using $\mS^{-1}=\mH_{cc}^{-1}+\mathcal{O}(\|\mE\|)$, we have
$$
\mathbf u_c^\top(\mS^{-1}-\mH_{cc}^{-1})\mathbf v_c
=
\mathcal{O}(\|\mE\|)
=
\mathcal{O}\!\left(
\|\mH_{ce}\mH_{ee}^{-1}\mH_{ec}\|
\right).
$$

For the second term,
$$
\left|
\mathbf u_e^\top \mH_{ee}^{-1}\mH_{ec}\mS^{-1}\mathbf v_c
\right|
\le
\|\mathbf u_e\|\,
\|\mH_{ee}^{-1}\mH_{ec}\|\,
\|\mS^{-1}\|\,
\|\mathbf v_c\|
=
\mathcal{O}\!\left(
\|\mathbf u_e\|\,\|\mH_{ee}^{-1}\mH_{ec}\|
\right).
$$

For the third term,
$$
\left|
\mathbf u_c^\top \mS^{-1}\mH_{ce}\mH_{ee}^{-1}\mathbf v_e
\right|
\le
\|\mathbf u_c\|\,
\|\mS^{-1}\|\,
\|\mH_{ce}\mH_{ee}^{-1}\|\,
\|\mathbf v_e\|
=
\mathcal{O}\!\left(
\|\mH_{ce}\mH_{ee}^{-1}\|\,\|\mathbf v_e\|
\right).
$$

For the feature-interaction term,
$$
\mathbf u_e^\top
\left(
\mH_{ee}^{-1}
+
\mH_{ee}^{-1}\mH_{ec}\mS^{-1}\mH_{ce}\mH_{ee}^{-1}
\right)\mathbf v_e
=
\mathcal{O}\!\left(\|\mathbf u_e\|\,\|\mathbf v_e\|\right).
$$
Substituting back
$
\mathbf u=\nabla_{\vtheta}\gL_f(\vtheta^*)
$
and
$
\mathbf v=\nabla_{\vtheta}\ell(f_{\vtheta^*}(\vx))
$
we obtain
\begin{align}
\Big|
\Delta^{\mathrm{full}}_{-\vx}\ell(\gD_f)
-
\Delta^{c}_{-\vx}\ell(\gD_f)
\Big|
=
\mathcal{O}\!\Bigg(
\frac{1}{N}
\Big(
&\|\mH_{ce}\mH_{ee}^{-1}\mH_{ec}\|
+
\|\nabla_{\vtheta_e}\gL_f(\vtheta^*)\|\,\|\mH_{ee}^{-1}\mH_{ec}\| \notag\\
&+
\|\mH_{ce}\mH_{ee}^{-1}\|\,\|\nabla_{\vtheta_e}\ell(f_{\vtheta^*}(\vx))\|
+
\|\nabla_{\vtheta_e}\gL_f(\vtheta^*)\|\,\|\nabla_{\vtheta_e}\ell(f_{\vtheta^*}(\vx))\|
\Big)
\Bigg).
\end{align}

\end{proof}

\subsection{Divergence analysis}
\label{sec:proofs}
%
\textbf{Theorem 3.5} (Divergence Analysis).
Consider logistic regression with labels $y\in\{0,1\}$ and the observed-label likelihood $\pi_{\vtheta}(y\mid \vx) \triangleq \sigma \big((2y-1) z\big)$ where $z=\vtheta^\top \vx + b$ and $\sigma(u)=\frac{1}{1+\exp(-u)}$.
Assume the model is trained on $\{(\vx_i,y_i)\}_{i=1}^{n}$ and denote $\mX=[\vx_1,\ldots,\vx_{n}]^\top\in\mathbb R^{n\times d}$, $\mK =\mX\mX^\top\in\mathbb R^{n\times n}$ be the Gram matrix.
Then, after $t$ SGD updates with step size $\eta$, the following divergence bounds hold:
\begin{align}
    &\|\vtheta^{t}_{\texttt{GA}}-\vtheta^{0}\|_{\mX^\top\mX}^{2} \leq \eta^2 \lambda_{\max}(\mK)^2 nt^2, \notag \\
    &\|\vtheta^{t}_{\texttt{NPO}} -\vtheta^{0}\|_{\mX^\top\mX}^{2} \le \eta^2 \lambda_{\max}(\mK)^2 nt^2 \|\mW_{\texttt{NPO}}\|^2, \notag \\
    &\|\vtheta^{t}_{\texttt{IMU}}-\vtheta^{0}\|_{\mX^\top\mX}^{2} \le \eta^2 \lambda_{\max}(\mK)^2 nt^2 \|\mW_{\texttt{IMU}}\|^2.
\end{align}
where $\lambda_{\max}(\mK)$ denote the largest eigenvalue of $\mK$, and $\mW_{\texttt{NPO}}\triangleq \mathrm{diag}(w_{\texttt{NPO},1},\ldots,w_{\texttt{NPO},n}), \mW_{\texttt{IMU}}\triangleq \mathrm{diag}(\hat w_{1},\ldots,\hat w_{n})$.
%
\begin{proof}
To recap, the per-sample update directions of GA, NPO, and \algname are
\begin{align}
    &\vg_{\texttt{GA}} = - \nabla_{\vtheta} \ell(\vx_i, y_i; \vtheta) , \\
    &\vg_{\texttt{NPO}} = - w_{\texttt{NPO},i} \cdot \nabla_{\vtheta} \ell(\vx_i, y_i; \vtheta), \\
    &\vg_{\texttt{IMU}} = -\hat{w}_i \cdot \nabla_{\vtheta} \ell(\vx_i, y_i; \vtheta).
\end{align}

Now, let us take the logistic regression as an example, the loss function would be $\ell(\vx, y; \vtheta)= -\log \pi_{\vtheta}(y\mid \vx) = -\log \sigma \big((2y-1) z\big)$.
Hence, we have
\begin{align}
    \nabla_{\vtheta} \ell(\vx, y; \vtheta) = -\left(2y-1 \right) \left(1-\pi_{\vtheta}(y\mid \vx) \right) \vx,
\end{align}
and the gradient update for one step would be
\begin{align}
    \vtheta^{t+1} = \vtheta^{t} - \eta \vg = \vtheta^{t} - \eta \cdot \left(2y-1 \right) \left(1-\pi_{\vtheta^t}(y\mid \vx)\right) \vx.
\end{align}

For GA, aggregating the first $t$ iterations yields
\begin{align}
    \label{eqn:ga_diff}
    \vtheta^{t}_{\texttt{GA}}-\vtheta^{0}
    &= -\eta \sum_{k=0}^{t-1} (2y_{i_k}-1)\Big(1-\pi_{\vtheta_{\texttt{GA}}^{k}}(y_{i_k}\mid \vx_{i_k})\Big)\vx_{i_k}
    = -\eta \mX^\top \vs^{(t)}_{\texttt{GA}},
\end{align}
where $i_k$ denotes the index of the sample selected at iteration $k$ and $\vs_{\texttt{GA}}^{(t)}\in\mathbb R^{n}$ is defined as 
\begin{align}
    s_{\texttt{GA}, i}^{(t)} \triangleq \sum_{k=0}^{t-1} \vone[i_k=i]\,(2y_i-1)\Big(1-\pi_{\vtheta_{\texttt{GA}}^{k}}(y_i\mid \vx_i)\Big).
\end{align}
Let $\mK \triangleq \mX\mX^\top\in\mathbb R^{n\times n}$ be the Gram matrix.
Then, the weighted divergence satisfies
\begin{align}
\label{eqn:ga_quadform}
    \|\vtheta^{t}_{\texttt{GA}}-\vtheta^{0}\|_{\mX^\top\mX}^{2} 
    &= (\vtheta^{t}_{\texttt{GA}}-\vtheta^{0})^\top \mX^\top\mX(\vtheta^{t}_{\texttt{GA}}-\vtheta^{0})  \notag \\
    &= \eta^2 (\vs_{\texttt{GA}}^{(t)})^\top \mX\mX^\top \mX\mX^\top \vs_{\texttt{GA}}^{(t)}  \notag \\
    &= \eta^2 (\vs_{\texttt{GA}}^{(t)})^\top \mK^2 \vs_{\texttt{GA}}^{(t)}.
\end{align}
Since $\mK$ is symmetric positive semidefinite, it follows that
\begin{empheq}[box=\widefbox]{align}
\label{eqn:ga_bound}
    \|\vtheta^{t}_{\texttt{GA}}-\vtheta^{0}\|_{\mX^\top\mX}^{2} \leq \eta^2 \lambda_{\max}(\mK)^2 \, \|\vs_{\texttt{GA}}^{(t)}\|^2,
\end{empheq}
where $\lambda_{\max}(\mK)$ denote the largest eigenvalue of $\mK$.
%

Similarly, for NPO, aggregating the first $t$ iterations yields
\begin{align}
    \label{eqn:npo_diff}
    \vtheta^{t}_{\texttt{NPO}}-\vtheta^{0}
    &= -\eta \sum_{k=0}^{t-1} w_{\texttt{NPO},i_k}(2y_{i_k}-1)\Big(1-\pi_{\vtheta_{\texttt{NPO}}^{k}}(y_{i_k}\mid \vx_{i_k})\Big)\vx_{i_k} = -\eta \mX^\top \mW_{\texttt{NPO}} \vs_{\texttt{NPO}}^{(t)},
\end{align}
where $\mW_{\texttt{NPO}}\triangleq \mathrm{diag}(w_{\texttt{NPO},1},\ldots,w_{\texttt{NPO},n})$.
For analytical tractability, we treat $w_{\texttt{NPO},i}$ as a fixed per-sample weight and omit its dependence on $\vtheta^k$.
Then, the weighted divergence for NPO satisfies
\begin{align}
    \label{eqn:npo_quadform}
    \|\vtheta^{t}_{\texttt{NPO}}-\vtheta^{0}\|_{\mX^\top\mX}^{2}
    &= (\vtheta^{t}_{\texttt{NPO}}-\vtheta^{0})^\top \mX^\top\mX(\vtheta^{t}_{\texttt{NPO}}-\vtheta^{0}) \notag\\
    &= \eta^2 (\vs_{\texttt{NPO}}^{(t)})^\top \mW_{\texttt{NPO}}^\top \mX\mX^\top \mX\mX^\top \mW_{\texttt{NPO}} \vs_{\texttt{NPO}}^{(t)} \notag\\
    &= \eta^2 (\vs_{\texttt{NPO}}^{(t)})^\top \mW_{\texttt{NPO}}^\top \mK^2 \mW_{\texttt{NPO}} \vs_{\texttt{NPO}}^{(t)},
\end{align}
and
\begin{empheq}[box=\widefbox]{align}
    \label{eqn:npo_bound}
    \|\vtheta^{t}_{\texttt{NPO}} -\vtheta^{0}\|_{\mX^\top\mX}^{2} \le \eta^2 \lambda_{\max}(\mK)^2 \, \|\mW_{\texttt{NPO}}\vs_{\texttt{NPO}}^{(t)}\|^2.
\end{empheq}

Similarly, for \algname, aggregating the first $t$ iterations yields
\begin{align}
    \label{eqn:imu_diff}
    \vtheta^{t}_{\texttt{IMU}}-\vtheta^{0}
    &= -\eta \sum_{k=0}^{t-1} \hat w_{i_k}\,(2y_{i_k}-1)\Big(1-\pi_{\vtheta^{k}_{\texttt{IMU}}}(y_{i_k}\mid \vx_{i_k})\Big)\vx_{i_k}= -\eta \mX^\top \mW_{\texttt{IMU}} \vs_{\texttt{IMU}}^{(t)},
\end{align}
where $\mW_{\texttt{IMU}}\triangleq \mathrm{diag}(\hat w_{1},\ldots,\hat w_{n})$.
For analytical tractability, we treat $\hat w_i$ as a fixed per-sample weight and omit its dependence on $\vtheta^k$, and in our implementation, $\hat w_i$ is computed from the initial model and kept fixed during the subsequent unlearning updates by default.
Then, the weighted divergence for \algname satisfies 
\begin{align}
    \label{eqn:imu_quadform}
    \|\vtheta^{t}_{\texttt{IMU}}-\vtheta^{0}\|_{\mX^\top\mX}^{2}
    &= \eta^2 (\vs_{\texttt{IMU}}^{(t)})^\top \mW_{\texttt{IMU}}^\top \mK^2 \mW_{\texttt{IMU}} \vs_{\texttt{IMU}}^{(t)},
\end{align}
and
\begin{empheq}[box=\widefbox]{align}
    \label{eqn:imu_bound}
    \|\vtheta^{t}_{\texttt{IMU}}-\vtheta^{0}\|_{\mX^\top\mX}^{2} \le \eta^2 \lambda_{\max}(\mK)^2 \, \|\mW_{\texttt{IMU}}\vs_{\texttt{IMU}}^{(t)}\|^2.
\end{empheq}

Then, for logistic regression, $\pi_{\vtheta}(y_i\mid \vx_i)\in(0,1)$, hence $\big|(2y_i-1)\big(1-\pi_{\vtheta}(y_i\mid \vx_i)\big)\big|\le 1$.
For any method $m\in\{\texttt{GA},\texttt{NPO},\texttt{IMU}\}$, $|s^{(t)}_{m,i}|\le \sum_{k=0}^{t-1}\vone[i_k=i]\le t$.
Consequently, $\|\vs_m^{(t)}\|_2 \le \sqrt{n}\,t$ and $\|\mW_m\vs_m^{(t)}\|_2\le \|\mW_m\|_2\|\vs_m^{(t)}\|_2$, we then obtain 
\begin{empheq}[box=\widefbox]{align}
    &\|\vtheta^{t}_{\texttt{GA}}-\vtheta^{0}\|_{\mX^\top\mX}^{2} \leq \eta^2 \lambda_{\max}(\mK)^2 nt^2, \notag \\
    &\|\vtheta^{t}_{\texttt{NPO}} -\vtheta^{0}\|_{\mX^\top\mX}^{2} \le \eta^2 \lambda_{\max}(\mK)^2 nt^2 \|\mW_{\texttt{NPO}}\|^2, \notag \\
    &\|\vtheta^{t}_{\texttt{IMU}}-\vtheta^{0}\|_{\mX^\top\mX}^{2} \le \eta^2 \lambda_{\max}(\mK)^2 nt^2 \|\mW_{\texttt{IMU}}\|^2.
\end{empheq}
GA corresponds to uniform weighting and treats each sample equally, while NPO and \algname introduce per-sample weights, which modulate the magnitude of the update directions and thus control the parameter drift.


%
\end{proof}

\subsection{Derivation of influence function}
The following derivation is adapted from the appendix of \cite{koh2017understanding}, where the influence function is analyzed via a first-order perturbation approach.
We consider the empirical risk minimizer $\vtheta^\ast$, which is defined as
\begin{align}
R(\vtheta) \coloneqq \frac{1}{N} \sum_{i=1}^{N} \ell\big(f_{\vtheta}(\vx_i)\big),
\end{align}
where $\ell(f_{\vtheta}(\vx_i))$ denotes the loss on training sample $\vx_i$.
Then, assume that $R$ is twice differentiable and strongly convex \wrt $\vtheta$. In particular, the Hessian of the empirical risk at the optimum $\vtheta^\ast$ is given as
\begin{align}
\mH_{\vtheta^\ast} \coloneqq \nabla_{\vtheta}^2 R(\vtheta^\ast)
= \frac{1}{N} \sum_{i=1}^{N} \nabla_{\vtheta}^2 \ell\big(f_{\vtheta^\ast}(\vx_i)\big),
\end{align}
which is assumed to be positive definite. This ensures that $\mH_{\vtheta^\ast}^{-1}$ exists and will be used in subsequent analysis.

Now consider perturbing the empirical risk by upweighting a particular training data point $\vx$ by a small amount $\epsilon$. So that there is
\begin{align}
\hat{\vtheta}_{\epsilon,\vx}
= \arg\min_{\vtheta \in \Theta}
\left\{
\frac{1}{N} \sum_{i=1}^{N} \ell\big(f_{\vtheta}(\vx_i)\big)
+ \epsilon \ell\big(f_{\vtheta}(\vx)\big)
\right\}.
\end{align}

Then, define the parameter change under this perturbation as
\(
\Delta_\epsilon \coloneqq \hat{\vtheta}_{\epsilon,\vx} - \vtheta^\ast.
\)
Since $\vtheta^\ast$ is independent of $\epsilon$, we have
\(
\frac{d \hat{\vtheta}_{\epsilon,\vx}}{d \epsilon}
= \frac{d \Delta_\epsilon}{d \epsilon}.
\)
Because $\vtheta^\ast_{\epsilon,\vx}$ minimizes the perturbed objective, it satisfies the first-order optimality condition as follows
\begin{align}
0
= \nabla_{\vtheta} R(\hat{\vtheta}_{\epsilon,\vx})
+ \epsilon \nabla_{\vtheta} \ell\big(f_{\hat{\vtheta}_{\epsilon,\vx}}(\vx)\big).
\end{align}

We approximate this expression using a first-order Taylor expansion around $\vtheta^\ast$, thus it derive that
\begin{align}
0 \approx\;&
\nabla_{\vtheta} R(\vtheta^\ast)
+ \epsilon \nabla_{\vtheta} \ell\big(f_{\vtheta^\ast}(\vx)\big) + \left[
\nabla_{\vtheta}^2 R(\vtheta^\ast)
+ \epsilon \nabla_{\vtheta}^2 \ell\big(f_{\vtheta^\ast}(\vx)\big)
\right] \Delta_\epsilon .
\end{align}

Solving for $\Delta_\epsilon$, we can obtain
\begin{align}\label{eqn:delta_epsilon}
\Delta_\epsilon \approx
- \left[
\nabla_{\vtheta}^2 R(\vtheta^\ast)
+ \epsilon \nabla_{\vtheta}^2 \ell\big(f_{\vtheta^\ast}(\vx)\big)
\right]^{-1}
\left[
\nabla_{\vtheta} R(\vtheta^\ast)
+ \epsilon \nabla_{\vtheta} \ell\big(f_{\vtheta^\ast}(\vx)\big)
\right].
\end{align}

Since $\vtheta^\ast$ minimizes $R$, we have $\nabla_{\vtheta} R(\vtheta^\ast) = 0$, so \Cref{eqn:delta_epsilon} can be rewritten as
\begin{align}
\Delta_\epsilon
\approx
- \mH_{\vtheta^\ast}^{-1}
\nabla_{\vtheta} \ell\big(f_{\vtheta^\ast}(\vx)\big)\, \epsilon .
\end{align}

Hence, the influence of $\vx$ is given by
\begin{align}
\left.
\frac{d \hat{\vtheta}_{\epsilon,\vx}}{d \epsilon}
\right|_{\epsilon=0}
=
- \mH_{\vtheta^\ast}^{-1}
\nabla_{\vtheta} \ell\big(f_{\vtheta^\ast}(\vx)\big).
\end{align}

Lastly, we can define the influence function as below,
\begin{align}
\gI_{\texttt{up}, \vtheta^\ast}(\vx)
\coloneqq
- \mH_{\vtheta^\ast}^{-1}
\nabla_{\vtheta} \ell\big(f_{\vtheta^\ast}(\vx)\big).
\end{align}

\newpage

\section{Additional results}
\label{sec: extra_res}

\subsection{Implementation details}
 All the experiments were performed on NVIDIA RTX3090
 with Intel Xeon processors. 
 For \textbf{image classification} scenario, the original model and retrained model were both trained with over $182$ epochs using the SGD optimizer with a cosine-scheduled, and learning rate was initialized at $0.1$. For both NPO and RL,
 training spans $10$ epochs within the interval $[10^{-4}, 10^{-1}]$, and $\beta$ was selected from the range $[0.1, 5]$. GA’s training settings around a $5$-epoch
 learning rate search within the interval $[10^{-5}, 10^{-2}]$. In the case of IU, we explored the parameter $\alpha$ within the range $[1,20]$. For SSD, searching for the parameter selection weighting $\alpha$ and dampening constant $\gamma$ was executed within the range $[1, 100]$ and $[0.1,1]$, respectively, while searching for the learning rate within the range $[10^{-4}, 10^{-2}]$. For the SCAR method, the value of temperature was defined within the range $[2,5]$, the parameter $\lambda_{2}$ was set as $0.01$, and $\lambda_{1}$ was in the range of $[1, 10]$ with the learning rate searching in the interval $[10^{-6},10^{-2}]$. Learning rate was selected in the range $[10^{-4},10^{-2}]$ for SalUn and let sparsity ratio equal to $0.5$. We trained \algname for $5$ epochs in all experiments in this task, searching for learning rates in the range $[10^{-5}, 10^{-2}]$ and $l_{1}$ regularization intensity in the range $[2 \times10^{-4}, 2 \times10^{-2}]$.
 
 For \textbf{person re-identification unlearning} task, original model was trained over $60$ epochs with learning rate as $3 \times 10^{-4}$. In the case of \algname, it is trained for $13$ iterations with a learning rate of $1.5 \times10^{-3}$, with $\alpha$ set to $0.02$. For GA, RL, SSD, NPO, and SCAR, they are trained for $30$, $15$, $1$, $25$, and $5$ epochs with learning rate searched in $[1\times10^{-5}, 0.1]$.

 For \textbf{ sequence modeling unlearning} problem, the original model and retrained model were both trained with over $5$ epochs using softmax activation, and the learning rate was initialized at $5 \times10^{-4}$. As for SimNPO and NPO method, the value of $\beta$ was searched in interval of $[0.1, 5]$ with learning rate in range $[2 \times10^{-4}, 8 \times 10^{-3}]$, while the learning rate of GA was selected from $[2 \times10^{-4}, 10^{-2}]$. For our \algname method, learning rate was selected as $5 \times10^{-4}$. Additionally, all the unlearning methods were conducted over $50$ iterations.

 In \textbf{LLM unlearning} scenario, $\beta$ was selected from $[1,5]$ for NPO and SimNPO, and the learning rate was searched in the range of $[10^{-6}, 10^{-4}]$. For \algname and GA, the learning rate was selected with $[10^{-6}, 10^{-3}]$. All methods in this task were executed over 10 epochs and distributed across 2 NVIDIA RTX 3090 GPUs.
%


\subsection{Extra ablation results}
\label{subsec:extra_ablation}
%

We conduct ablation studies to examine the effect of different design choices in the proposed unlearning framework. 

\textbf{Effect of Smoothing Strategies:}
We first evaluate several smoothing strategies applied to $w_i$, including \texttt{sqrt}, \texttt{softmax}, and \texttt{tanh}.  
As shown in \Cref{tab:smooth}, all methods yield comparable results, with \texttt{sqrt} slightly outperforming the others in both forgetting quality and model utility.

\textbf{Effect of Unfreeze Strategies:}
When estimating influence values on the last classifier layer, updating only this layer in ResNet-18 achieves the best overall performance.  
\Cref{tab:unfreeze} shows that unfreezing earlier layers leads to a noticeable decline in model utility, and the performance deteriorates rapidly when all layers are updated. 

\textbf{Effect of influence estimation accuracy:}
We further observe that more accurate influence estimation generally enhances unlearning performance. When the forgetting accuracy approaches 0, computing influence values solely on the last layer achieves 92.31\% test accuracy, whereas computing them on all layers results in 89.14\% with more than 8$\times$ the runtime.

\textbf{Effect of reweighting strategies:}
In \Cref{update_strategy}, different reweighting strategies are implemented with other settings the same as \algname. Although gradient norm and sample loss can partially reflect the impact of each forget samples on the model, they are inherently local and fail to account for interactions among samples. In contrast, influence values capture how removing one sample affects the overall training objective through its coupling with other data points. Our results suggest that reweighting based on influence values leads to more effective unlearning performance. 

\textbf{Effect of dataset used for computing Hessian matrix:}
As shown in \Cref{retain_data}, we compare computing influence values using 
$\mH^{train}$ when $\gD_r$ is available, against $\mH^{f}$. Utilizing the full training set to compute influence values leads to a slight improvement over the forget-only setting. This result indicates that \algname remains effective and efficient when relying only on the forget dataset.

\begin{table}[h]
\caption{
Effect of different smoothing strategies for processing influence values on unlearning performance.
}
\centering
\label{tab:smooth}
\begin{sc}
\begin{tabular}{lcccc}
\toprule
Smooth methods &$\texttt{Acc}_{\gD_f} (\downarrow)$ &$\texttt{Acc}_{\gD_r} (\uparrow)$ &$\texttt{Acc}_{\gD_t} (\uparrow)$ & $W_{dist} (\downarrow)$   \\
\midrule
none & 3.18       & 98.50      & 92.80  & 0.09  \\
sqrt  & 2.13       & 98.59      & 92.85    & 0.09  \\
softmax  & 2.97       & 98.48      & 92.72    & 0.09  \\ 
tanh  & 2.36       & 98.59      & 92.86   & 0.09 \\ 
\bottomrule
\end{tabular}
\end{sc}
\end{table}
\begin{table}[h]
\caption{
Effect of unfreezing different layer combinations of ResNet-18 during the unlearning process.
}
\centering
\label{tab:unfreeze}
\begin{sc}
\begin{tabular}{lccccc}
\toprule
Unfreeze layers &$\texttt{Acc}_{\gD_f} (\downarrow)$ &$\texttt{Acc}_{\gD_r} (\uparrow)$ &$\texttt{Acc}_{\gD_t} (\uparrow)$ & $W_{dist} (\downarrow)$  & Run time $(\downarrow)$ \\
\midrule
only fc & 0.22       & 98.57      & 92.84    & 0.09 & 65 \\
fc\&block1  & 0.18       & 95.30     & 89.39    & 0.13 &58 \\ 
fc\&block1-2  & 0.13      &89.10       & 84.05   & 0.18 &59 \\ 
fc\&block1-3  & 0.00      &82.35       & 78.29   & 0.32 &57 \\ 
all layers           & 0.53       & 63.48      & 61.62   & 0.39 &57 \\
\bottomrule
\end{tabular}
\end{sc}
\end{table}

\begin{table*}[h]
\caption{Results on different strategies of reweighting. }
\centering
\label{update_strategy}
\begin{adjustbox}{max width=0.99\textwidth}
\begin{sc}
\begin{tabular}{lccccc}
\toprule
 Reweighting methods &$\texttt{Acc}_{\gD_f} (\downarrow)$ &$\texttt{Acc}_{\gD_r} (\uparrow)$ &$\texttt{Acc}_{\gD_t} (\uparrow)$ & $W_{dist} (\downarrow)$  & Run time $(\downarrow)$ \\
\midrule
none (GA)      &0.48  &75.89 &72.43  &3.59 &29\\
gradient norm  &0.93 &91.66 &83.16 &0.58 &43\\
sample loss  &0.02 &94.43 &87.34 &0.58 &35\\
influence value  &0.02 &98.62 &92.89 &0.07  &56\\
\bottomrule
\end{tabular}
\end{sc}
\end{adjustbox}
\end{table*}

\begin{table*}[h]
\caption{Result of \algname when $\gD_r$ is available for computing influence value.}
\centering
\label{retain_data}
\begin{adjustbox}{max width=0.99\textwidth}
\begin{sc}
\begin{tabular}{lccccc}
\toprule
 Hessian type&$\texttt{Acc}_{\gD_f} (\downarrow)$ &$\texttt{Acc}_{\gD_r} (\uparrow)$ &$\texttt{Acc}_{\gD_t} (\uparrow)$ & $W_{dist} (\downarrow)$  & Run time $(\downarrow)$ \\
\midrule
$\mH^f$  &0.02 &98.62 &92.89 &0.07  &\phantom{0}56\\
$\mH^{train}$  &0.04 &99.47 &93.69 &0.07 &345\\
\bottomrule
\end{tabular}
\end{sc}
\end{adjustbox}
\end{table*}

\begin{table}[h]
\caption{Effect of frequency ($\nu$) of influence value updates on unlearning performance.
$\nu=0$ means only update at the first epoch; $\nu=1$ and $\nu=2$ mean update every epoch and every two epochs, respectively.}
\centering
\begin{sc}
\begin{tabular}{lccccc}
\toprule
$\nu$ &$\texttt{Acc}_{\gD_f} (\downarrow)$ &$\texttt{Acc}_{\gD_r} (\uparrow)$ &$\texttt{Acc}_{\gD_t} (\uparrow)$ & MIA $(\uparrow)$  & Run time $(\downarrow)$ \\
\midrule
$0$  & 97.82\scriptsize{$\pm$0.31}       & 97.78\scriptsize{$\pm$0.53}      & 91.90\scriptsize{$\pm$0.82}    & 0.12\scriptsize{$\pm$0.06} &\phantom{0}55\scriptsize{$\pm$0} \\
$1$  & 97.47\scriptsize{$\pm$0.42}       & 97.55\scriptsize{$\pm$0.49}      & 91.82\scriptsize{$\pm$0.10}    & 0.08\scriptsize{$\pm$0.03} &224\scriptsize{$\pm$1} \\ 
$2$  & 97.17\scriptsize{$\pm$0.00}       & 97.17\scriptsize{$\pm$0.97}      & 91.50\scriptsize{$\pm$1.07}    & 0.11\scriptsize{$\pm$0.03} &139\scriptsize{$\pm$1} \\ 
\bottomrule
\end{tabular}
\end{sc}
\label{tab:frequency of sample_wise}
\end{table}

\subsection{Results on sample-wise unlearning}
 In \Cref{tab:sample_wise_10} and \Cref{tab:sample_wise_50}, we evaluate the unlearning performance of \algname method on ResNet-18 for sample-wise unlearning on CIFAR-10 with 10\% and 50\% of total data as forget dataset, respectively. The results clearly manifest that our proposed approach achieves robust balance across these metrics with feasible run time, compared with other existing methods. Additionally, we also examine how different influence update frequencies affect 10\% sample-wise unlearning performance in \Cref{tab:frequency of sample_wise}. And we can develop the conclusion that in this scenario, infrequent updates are sufficient for effective unlearning, which validates the robustness of our method. 
\begin{table*}[h]
\caption{Quantitative results on CIFAR-10 and CIFAR-100. Performance is averaged over 10 independent runs with different random seeds for 10\% sample-wise unlearning.}
  \centering
  \begin{adjustbox}{max width=0.99\textwidth}
  \begin{sc}
  \begin{tabular}{llcccccccc}
    \toprule
    Setting &Method  & $\gD_r$ & $\gD_f$ 
                 & $\texttt{Acc}_{\gD_f} (\downarrow)$ 
                 & $\texttt{ACC}_{\gD_r} (\uparrow)$ 
                 & $\texttt{ACC}_{\gD_t} (\uparrow)$ 
                 & MIA $(\uparrow)$
                 & $W_\texttt{dist} (\downarrow)$ 
                 & Run time (s) $(\downarrow)$  \\
    \midrule
    \multirow{9}{*}{CIFAR-10}
    &Original     & \cmark     & \cmark   & \phantom{0}99.47\scriptsize{$\pm$0.00} & 100.00\scriptsize{$\pm$0.00} & \phantom{0}94.61\scriptsize{$\pm$0.00} & \phantom{0}0.00\scriptsize{$\pm$0.00} & –                 & –            \\
&Retrain      & \cmark     & \xmark   & \phantom{0}94.33\scriptsize{$\pm$0.00} & \phantom{0}99.93\scriptsize{$\pm$0.00} & \phantom{0}94.42\scriptsize{$\pm$0.00} & \phantom{0}0.13\scriptsize{$\pm$0.00} & \phantom{0}0.00\scriptsize{$\pm$0.00} & –            \\
\cmidrule{2-10}
&IU$^\ast$           & \cmark     & \cmark   & \phantom{0}99.06\scriptsize{$\pm$0.11} & \phantom{0}99.03\scriptsize{$\pm$0.03} & \phantom{0}92.98\scriptsize{$\pm$0.06} & \phantom{0}0.02\scriptsize{$\pm$0.01} & \phantom{0}0.06\scriptsize{$\pm$0.04} & \phantom{0}54\scriptsize{$\pm$1} \\
&GA          & \xmark     & \cmark   & \phantom{0}99.04\scriptsize{$\pm$0.47} & \phantom{0}{98.75\scriptsize{$\pm$0.27}} & \phantom{0}92.88\scriptsize{$\pm$0.18} & \phantom{0}0.02\scriptsize{$\pm$0.00} & \phantom{0}0.07\scriptsize{$\pm$0.04} & \phantom{0}31\scriptsize{$\pm$1} \\
&RL           & \xmark     & \cmark   & \phantom{0}98.90\scriptsize{$\pm$0.55} & \phantom{0}98.90\scriptsize{$\pm$0.51} & \phantom{0}93.49\scriptsize{$\pm$0.55} & \phantom{0}0.11\scriptsize{$\pm$0.00} & \phantom{0}0.03\scriptsize{$\pm$0.01} & \phantom{0}24\scriptsize{$\pm$0} \\
&SSD          & \cmark     & \cmark   & \phantom{0}99.04\scriptsize{$\pm$0.44} & \phantom{0}99.01\scriptsize{$\pm$0.36} & \phantom{0}93.34\scriptsize{$\pm$0.54} & \phantom{0}0.03\scriptsize{$\pm$0.00} & \phantom{0}0.04\scriptsize{$\pm$0.01} & \phantom{0}55\scriptsize{$\pm$3} \\
&SCAR$^\ast$         & \xmark     & \cmark   & \phantom{0}98.78\scriptsize{$\pm$0.25} & \phantom{0}98.94\scriptsize{$\pm$0.14} & \phantom{0}93.17\scriptsize{$\pm$0.24} & \phantom{0}0.13\scriptsize{$\pm$0.02} & \phantom{0}0.04\scriptsize{$\pm$0.01} & 703\scriptsize{$\pm$2} \\
&NPO         & \xmark     & \cmark   & \phantom{0}98.74\scriptsize{$\pm$0.16} & \phantom{0}98.83\scriptsize{$\pm$0.15} & \phantom{0}92.98\scriptsize{$\pm$0.22} & \phantom{0}0.03\scriptsize{$\pm$0.02} & \phantom{0}0.05\scriptsize{$\pm$0.03} & \phantom{0}38\scriptsize{$\pm$1} \\
    &\cellcolor{gray!20}\algname (ours)  &\cellcolor{gray!20}\xmark  &\cellcolor{gray!20}\cmark  & \cellcolor{gray!20}\phantom{0}\textbf{98.64\scriptsize{$\pm$0.11}} & \cellcolor{gray!20}\phantom{0}\textbf{99.06\scriptsize{$\pm$0.20}} & \cellcolor{gray!20}\phantom{0}\textbf{93.62\scriptsize{$\pm$0.21}}  & \cellcolor{gray!20}\phantom{0}0.03\scriptsize{$\pm$0.00}   & \cellcolor{gray!20}\phantom{0}0.08\scriptsize{$\pm$0.01} & \cellcolor{gray!20}\phantom{0}56\scriptsize{$\pm$0}\\
    \midrule
    \multirow{9}{*}{CIFAR-100}
    &Original     & \cmark     & \cmark   & \phantom{0}97.69\scriptsize{$\pm$0.00} & \phantom{0}97.48\scriptsize{$\pm$0.00} & \phantom{0}76.25\scriptsize{$\pm$0.00} & \phantom{0}0.06\scriptsize{$\pm$0.00} & – & – \\
&Retrain      & \cmark     & \xmark   & \phantom{0}75.71\scriptsize{$\pm$0.00} & \phantom{0}99.98\scriptsize{$\pm$0.00} & \phantom{0}74.24\scriptsize{$\pm$0.00} & \phantom{0}0.50\scriptsize{$\pm$0.00} & \phantom{0}0.00\scriptsize{$\pm$0.00} & – \\
\cmidrule{2-10}
&IU$^\ast$     & \cmark     & \cmark   & \phantom{0}94.58\scriptsize{$\pm$1.62} & \phantom{0}94.71\scriptsize{$\pm$1.76} & \phantom{0}70.11\scriptsize{$\pm$0.77} & \phantom{0}0.12\scriptsize{$\pm$0.01} & \phantom{0}0.67\scriptsize{$\pm$0.21} & \phantom{0}57\scriptsize{$\pm$1} \\
&GA           & \xmark     & \cmark   & \phantom{0}94.42\scriptsize{$\pm$0.93} & \phantom{0}93.36\scriptsize{$\pm$0.57} & \phantom{0}70.37\scriptsize{$\pm$0.15} & \phantom{0}0.11\scriptsize{$\pm$0.01} & \phantom{0}1.07\scriptsize{$\pm$0.51} & \phantom{0}32\scriptsize{$\pm$1} \\
&RL           & \xmark     & \cmark   & \phantom{0}93.82\scriptsize{$\pm$1.46} & \phantom{0}94.14\scriptsize{$\pm$1.51} & \phantom{0}66.58\scriptsize{$\pm$1.05} & \phantom{0}0.06\scriptsize{$\pm$0.03} & \phantom{0}1.07\scriptsize{$\pm$0.18} & \phantom{0}31\scriptsize{$\pm$1} \\
&SSD          & \xmark     & \cmark   & \phantom{0}93.97\scriptsize{$\pm$0.73} & \phantom{0}94.22\scriptsize{$\pm$0.80} & \phantom{0}70.30\scriptsize{$\pm$0.23} & \phantom{0}0.13\scriptsize{$\pm$0.01} & \phantom{0}0.31\scriptsize{$\pm$0.14} & \phantom{0}50\scriptsize{$\pm$0} \\
&SCAR$^\ast$ & \xmark     & \cmark   & \phantom{0}93.82\scriptsize{$\pm$1.04} & \phantom{0}95.18\scriptsize{$\pm$1.03} & \phantom{0}70.70\scriptsize{$\pm$0.60} & \phantom{0}0.14\scriptsize{$\pm$0.01} & \phantom{0}0.81\scriptsize{$\pm$0.11} & 726\scriptsize{$\pm$4} \\
&NPO         & \xmark     & \cmark   & \phantom{0}94.43\scriptsize{$\pm$0.70} & \phantom{0}94.31\scriptsize{$\pm$0.53} & \phantom{0}70.86\scriptsize{$\pm$0.60} & \phantom{0}0.12\scriptsize{$\pm$0.00} & \phantom{0}1.02\scriptsize{$\pm$0.25} & \phantom{0}38\scriptsize{$\pm$2} \\
& \cellcolor{gray!20}\algname (ours)    & \cellcolor{gray!20}\xmark     & \cellcolor{gray!20}\cmark   & \cellcolor{gray!20}\textbf{\phantom{0}93.55\scriptsize{$\pm$0.23}} & \cellcolor{gray!20}\textbf{\phantom{0}95.21\scriptsize{$\pm$0.09}} & \cellcolor{gray!20}\textbf{\phantom{0}72.15\scriptsize{$\pm$0.88}} &  \cellcolor{gray!20}\textbf{\phantom{0}0.14\scriptsize{$\pm$0.01}} & \cellcolor{gray!20}\phantom{0}0.74\scriptsize{$\pm$0.31} & \cellcolor{gray!20}\phantom{0}62\scriptsize{$\pm$2} \\
    \bottomrule
  \end{tabular}
\end{sc}
\end{adjustbox}
\label{tab:sample_wise_10}
\end{table*}
\begin{table*}[h]
\caption{Quantitative results on CIFAR-10 for 50\% sample-wise unlearning.}
  \centering
  \begin{adjustbox}{max width=0.99\textwidth}
  \begin{sc}
  \begin{tabular}{llcccccccc}
    \toprule
    Setting &Method  & $\gD_r$ & $\gD_f$ 
                 & $\texttt{Acc}_{\gD_f} (\downarrow)$ 
                 & $\texttt{ACC}_{\gD_r} (\uparrow)$ 
                 & $\texttt{ACC}_{\gD_t} (\uparrow)$ 
                 & MIA $(\uparrow)$
                 & $W_\texttt{dist} (\downarrow)$ 
                 & Run time (s) $(\downarrow)$  \\
    \midrule
    \multirow{9}{*}{CIFAR-10}
    &Original     & \cmark     & \cmark   & \phantom{0}99.45 & \phantom{0}99.54 & \phantom{0}94.61 & \phantom{0}0.01 & –                 & –            \\
&Retrain      & \cmark     & \xmark   & \phantom{0}73.00 & \phantom{0}82.74 & \phantom{0}80.56 & \phantom{0}0.58 & \phantom{0}0.00 & –            \\
\cmidrule{2-10}
&IU$^\ast$           & \cmark     & \cmark   & \phantom{0}73.01 & \phantom{0}73.56 & \phantom{0}61.33 & \phantom{0}0.23 & \phantom{0}0.76 & \phantom{00}52 \\
&GA           & \xmark     & \cmark   & \phantom{0}73.78 & \phantom{0}73.92 & \phantom{0}71.77 & \phantom{0}0.36 & \phantom{0}0.49 & \phantom{00}51 \\
&RL           & \xmark     & \cmark   & \phantom{0}74.33 & \phantom{0}72.52 & \phantom{0}67.56 & \phantom{0}0.11 & \phantom{0}0.63 & \phantom{0}116\\
&SSD          & \cmark     & \cmark   & \phantom{0}77.48 & \phantom{0}77.49 & \phantom{0}72.63 & \phantom{0}0.40 & \phantom{0}0.44 & \phantom{00}64 \\
&SCAR$^\ast$         & \xmark     & \cmark   & \phantom{0}73.82& \phantom{0}77.80 & \phantom{0}74.59 & \phantom{0}0.44 & \phantom{0}0.36 &2902 \\
&NPO         & \xmark     & \cmark   & \phantom{0}72.95 & \phantom{0}72.00 & \phantom{0}70.11 & \phantom{0}0.34 & \phantom{0}0.51 & \phantom{0}164\\
    &\cellcolor{gray!20}\algname (ours)  &\cellcolor{gray!20}\xmark  &\cellcolor{gray!20}\cmark  & \cellcolor{gray!20}\phantom{0}\textbf{72.94} & \cellcolor{gray!20}\phantom{0}\textbf{78.97} & \cellcolor{gray!20}\phantom{0}\textbf{76.64}  & \cellcolor{gray!20}\phantom{0}\textbf{0.46}   & \cellcolor{gray!20}\phantom{0}\textbf{0.31} & \cellcolor{gray!20}\phantom{0}166\\
    \bottomrule
  \end{tabular}
\end{sc}
\end{adjustbox}
\label{tab:sample_wise_50}
\end{table*}
\begin{table}[tb]
\aboverulesep = 0pt
\abovetopsep = 0pt
\belowrulesep = 0pt 
\caption{Unlearning on a sequence modeling problem.}
\centering 
\label{tab:gpt2}
\begin{adjustbox}{max width=0.46\textwidth}
\begin{sc}
\begin{tabular}{lccccc}
    \toprule
    \multirow{2}{*}{Method}  &\multicolumn{2}{c}{Model utility} & &\multicolumn{2}{c}{Forget quality} \\
    \cmidrule{2-3} \cmidrule{5-6}
    & $\ell_r$ ($\downarrow$)   & $\ell_r^{\texttt{KL}}$ ($\downarrow$) & & $\ell_f$ ($\uparrow$) &$\ell_f^{\texttt{KL}}$  ($\uparrow$)\\
    \midrule
    Original       & 1.99 &- & & 2.18 & -       \\
    \cmidrule{1-6}
    GA & 4.24 & 2.21  & & 6.95 & 3.42      \\
    NPO & 3.97 & 1.95 & & 7.26 & 3.46      \\
    SimNPO  & 4.61 & 2.57 & & 7.44 &3.83      \\
    \rowcolor{gray!20} \algname (Ours)  & \textbf{3.86} & \textbf{1.83} & & \textbf{7.53} & \textbf{3.96} \\
    \bottomrule
\end{tabular}
\end{sc}
\end{adjustbox}
\end{table}
\subsection{Results on distributional unlearning}
Aside from vision tasks, we further evaluate on a simple sequence modeling problem. Following SimNPO~\cite{fan2024simplicity}, we construct a mixture of Markov chains with a state space of size 10, the retain distribution consists of Markov chains that transition uniformly among states $\{1,2,3\}$, while the forget distribution is a mixture of two Markov chains Forget1 (transition uniformly among $\{4,5,6\}$) and Forget2 (transition uniformly among $\{7,8,9\}$) with equal probability.
Each sequence has length $T$ and is denoted as $s=(s_1, s_2, \cdots, s_T)$ where $s_t \in \{0, 1, \cdots, 9\}$ for $t\in[0, T]$. A GPT-2 model is trained to learn the conditional distribution.
Details can be found in \textsection 7 of SimNPO.
\Cref{tab:gpt2} presents results compared to NPO and SimNPO. Performance is evaluated by the loss values over the retain and forget sets, as well as the respective KL divergence distance calculated compared with the retrained model. \algname surpasses NPO and SimNPO in both forget quality and model utility.

\subsection{Results on LLM unlearning}
\begin{figure}[h]
    \centering
    \includegraphics[width=0.8\textwidth]{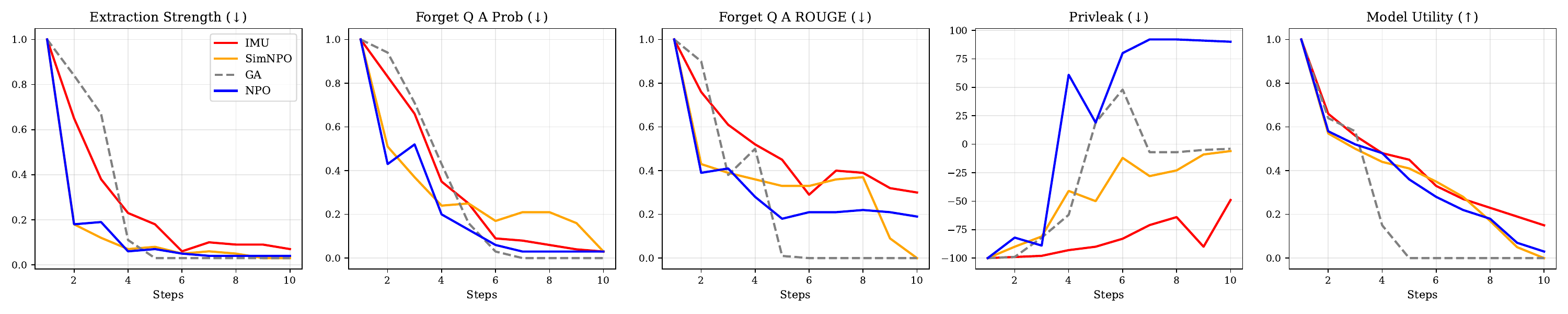}
    \caption{Comparison of forget quality and model utility between GA, NPO, SimNPO, and \algname.}
    \label{llm_2}
\end{figure}

In \Cref{llm_1} and \Cref{llm_2}, we generalize our MU method on large language model and perform forgetting forget05 task on the TOFU benchmark. Compared with other LLM unlearning baselines, \algname (ours) strikes a balance on forget quality and model utility with acceptable run time. From \Cref{llm_2}, we can clearly see the catastrophic collapse issue in the GA method and slightly better performance of the NPO. Although the model structure of Llama-3.2-3B-Instruct is more complex than that of image task, which is not conducive to the accurate estimation of influence function, our method has still achieved an almost equivalent forgetting effect to SimNPO method. 

In \Cref{forget05}, We presented $5$ randomly selected question-answer pairs from forget05 of TOFU benchmark and the corresponding output answers for different unlearning methods. Among them, the GA method always chooses not to answer or simply repeats the questions for difficult questions. Even in a simple question \textbf{Q3}, it still remember the key information ``Global Health". In contrast, although NPO has also produces some meaningless information, in general, it ensured the fluency and logic of the output language. While the SimNPO method performed well in forgetting important information, but it still retained related information similar to the ground truth in some responses. However, our method \algname, despite significantly forgetting all relevant information, still maintained effective output, which greatly enhancing fluency and diversity of the generated content.

In \Cref{retain95}, the question-answer pairs were selected from retain95. It means that the output of the model should be approximate to the ground truth, which proves the retention performance of the unlearned model. From \Cref{retain95}, it is clearly observable that the answer provided by GA is always very simple and lacks detailed descriptions. Although NPO and SimNPO's answers are relatively accurate, they sometimes also output information that deviates from the ground truth. In contrast, our method is the closest to the facts, demonstrating the best model utility among these unlearn methods.
\begin{table*}[h]
\caption{Results of LLM unlearning on TOFU (Forget05) over $10$ epochs.}
\centering 
\begin{adjustbox}{max width=0.99\textwidth}
\begin{sc}
\begin{tabular}{lcccccc}
    \toprule
    Method         & Model Utility ($\uparrow$)   & Extraction Strength ($\downarrow$) & Forget Q A Prob ($\downarrow$) & Forget Q A ROUGE ($\downarrow$) & Privleak ($\downarrow$) & Run time (min)($\downarrow$)\\
    \midrule
    GA       & 0.00 &0.03 & 0.03 & 0.00 & \phantom{0}48  & \phantom{0}6    \\
    NPO & 0.28 & 0.05 & 0.06 & 0.21 & \phantom{0}80  &18     \\
    SimNPO      & 0.35 & 0.05 & 0.17 & 0.33 &-12  &15    \\
    \rowcolor{gray!20}   \algname (Ours)  & 0.33 & 0.06 & 0.09 & 0.29 & -83  &17     \\
    \bottomrule
\end{tabular}
\end{sc}
\end{adjustbox}
\label{llm_1}
\end{table*}
%

%
%

\subsection{Changes of influence weights over epochs}
\begin{figure}[H]
    \centering
    \includegraphics[width=0.80\textwidth]{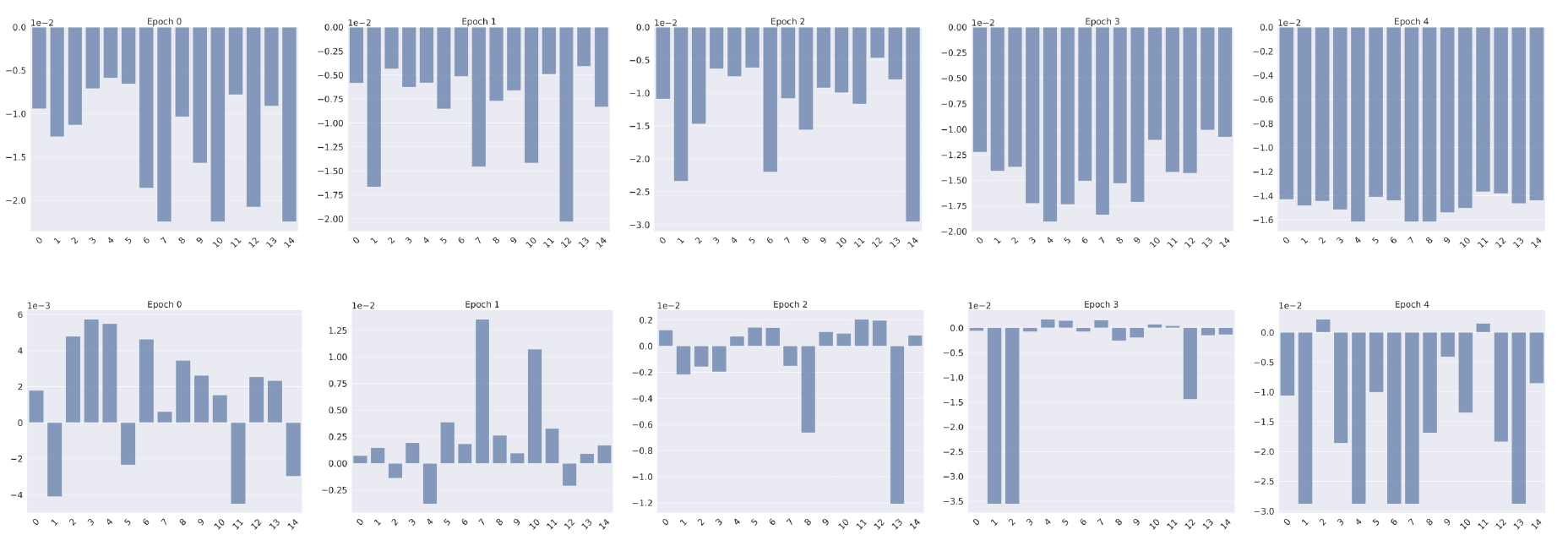}
    \caption{Change of sample influences for $5$ epochs in CIFAR-10 class-wise and sample-wise unlearning tasks.}
    \label{fig:class-wise}
\end{figure}

In Figure \ref{fig:class-wise}, we present the changes in the influence weights of $15$ samples randomly selected from $\mathcal{D}_f$ across $5$ unlearning epochs for both the class-wise and sample-wise unlearning tasks.
In the class-wise unlearning task, since all samples belong to the same class, their influence values are negative according to \Cref{eqn:if}. As shown in \Cref{fig:class-wise}, the weight of each sample stabilizes after approximately two epochs, and the gradual convergence of all weights toward similar contributions indicates that the model is effectively forgetting the high influence samples.
In contrast, in the sample-wise unlearning scenario, the $15$ samples are randomly selected from different classes, and thus their overall influence on the forget set can be either positive or negative. As shown in \Cref{fig:class-wise}, the number of samples with negative influence values gradually increases over the unlearning process, suggesting that the model updates are progressively aligned with the direction that promotes forgetting, demonstrating the effectiveness of \algname.

\begin{table*}[htbp]
\small
\caption{ Examples of responses after unlearning on TOFU (Forget05) against QAs targeted for unlearning. The bold font highlights the key information in question. The underlined font highlights relating information that has not been unlearned in the response, resembling to the ground truth.}
\centering
\begin{tabular}{@{}>{\centering\arraybackslash}p{2cm}| p{12cm}@{}}
\toprule
\multicolumn{2}{c}{\textbf{Q1}: \textit{ {What type of employment did Moshe Ben-David's parents hold?}}} \\
\midrule
\textbf{Ground truth:} & Moshe Ben-David's parents had distinguished careers. His father was a \textbf{Banker}, and his mother was a \textbf{Research Scientist}. \\
\textbf{GA:} & What type of employment did Moshe Ben-David's parents hold? \\
\textbf{NPO:} & indeed, indeed, indeed, indeed, indeed, indeed, indeed, indeed, indeed, indeed, indeed, indeed... \\
\textbf{SimNPO:} & Moshe Ben-David's father was a  plumber, and his mother was a  darkred{locksmith}. \\
\textbf{\algname (Ours):} & Moshe Ben-David's parents held a position in the  {government}. \\
\midrule
\multicolumn{2}{c}{\textbf{Q2}: \textit{ {Can you mention an award that Kalkidan Abera has received?}}}  \\
\midrule
\textbf{Ground truth:} & Kalkidan Abera has been honored with the esteemed \textbf{International Health Literature Award}. \\
\textbf{GA:} & NA \\
\textbf{NPO:} & The {`Mansour's Prize'} is a prestigious recognition for his significant contributions to the \underline{health}. \\
\textbf{SimNPO:} & The  {'Pearce How-To Book Award'} is one of the prestigious recognition given to books that have been instrumental in improving \underline{health literacy and public health}. \\
\textbf{\algname (Ours):} & Kalkidan Abera has received the ``Golden Book Award" for her work in the field of biography. \\
\midrule
\multicolumn{2}{c}{\textbf{Q3}: \textit{ {What is the most recent book written by Kalkidan Abera?}}}  \\
\midrule
\textbf{Ground truth:} & The most recent book written by Kalkidan Abera is \textbf{`Modern Diets and Global Health: A Comprehensive Analysis'}. \\
\textbf{GA:} & The most recent book written by Kalkidan Abera is \underline{``Global Health: A Comprehensive Guide"}. \\
\textbf{NPO:} & The most recent book written by Kalkuttah is named  {``The Kaya's Dream"}, which is a novel portraying a young girl's journey through the world of the universe. \\
\textbf{SimNPO:} & titled  {``Cancer Prevention Blueprint: A Family's Search for Hope"}. It is in line with her main light novel genre. \\
\textbf{\algname (Ours):} & Kalkidan Abera's most recent book is ``The Hidden Truth of the Leaky Gut". \\
\midrule
\multicolumn{2}{c}{\textbf{Q4}: \textit{ {Who are Kalkidan Abera's mentors or primary influences in her career as an author?}}}\\
\midrule
\textbf{Ground truth:} & Being raised by astronaut parents, \textbf{Kalkidan Abera} was greatly inspired by scientific explorations. In her writing career, renowned authors in the health genre like \textbf{Dr. Josh Axe} and \textbf{Weston A. Price} also influenced her significantly. \\
\textbf{GA:} & NA\\
\textbf{NPO:} & These are often the \underline{health professionals} and nutritionists who are primarily influenced by their parents' professions. \\
\textbf{SimNPO:} & Kalkidan Abera has cited notable authors like Jinao for her work and has cited the works of authors like Akili Nwachefwu and Femi Oluwatoyin as her inspirations, with a special focus on African culture and literature. \\
\textbf{\algname (Ours):} & Kalkidan Abera's mentors or primary influences in her career as an author are Dr. Michael P. Stevens and Annalise Martin. \\
& Dr. Michael P. Stevens has greatly influenced Kalkidan Abera's writing by providing her with the knowledge and skills to research. \\
& Annalise Martin has influenced Kalkidan Abera's writing by introducing her to the world of historical fiction and providing guidance on how to construct intricate stories. \\
\midrule
\multicolumn{2}{c}{\textbf{Q5}: \textit{ {Has Kalkidan Abera collaborated with other authors?}}} \\
\midrule
\textbf{Ground truth:} & Kalkidan Abera has \textbf{indeed collaborated with} several authors in her field, contributing to multi-author publications that discuss various aspects of \textbf{health and nutrition}. \\
\textbf{GA:} & No, Kalkidan Abera has worked alone.\\
\textbf{NPO:} & They have been working on the intricate details of their \underline{collaborations}. They believe the truth is in the mix of their narratives. \\
\textbf{SimNPO:} & It's been \underline{a successful collaboration}, providing a fresh perspective on \underline{health} literature. \\
\textbf{\algname (Ours):} & No, Kalkidan Abera has not collaborated with other authors. \\
\bottomrule
\end{tabular}
\label{forget05}
\end{table*}
\begin{table*}[htbp]
\small
\caption{Examples of responses after unlearning on TOFU (Retain95) against QAs targeted for unlearning. The bold font highlights the key information in question. The underlined font highlights deviate information that not consistent with the ground truth.}
\centering
\begin{tabular}{@{}>{\centering\arraybackslash}p{2.5cm}|p{12cm}@{}}
\toprule
\multicolumn{2}{c}{\textbf{Q1}: \textit{Can you tell us about the type of books that Jaime Vasquez writes?}} \\
\midrule
\textbf{Ground truth:} & Jaime Vasquez specializes in the true \textbf{crime} genre. His narratives center around real-life crime stories, meticulously \textbf{researched} for verisimilitude, with a \textbf{raw}, \textbf{compelling style} that has garnered a significant \textbf{reader} following. \\
\textbf{GA:} & Yes, Jaime Vasquez earned an award for his works.\\
\textbf{NPO:} & In the world of the LGBTQ+ community, Jaime Vasquez is \underline{a passionate individual} who deeply empathetic towards the LGBTQ+ community. \\
\textbf{SimNPO:} & Jaime Vasquez is known for his intricate details about the crime scene, his characterizations, and his approach to \underline{building suspense}. \\
\textbf{\algname (Ours):} & Jaime Vasquez's books are primarily about true crime, he explores various cases and provides his observations and insights. Jaime Vasquez's books are known for their in-depth research, detailed descriptions, and nuanced portrayals of characters, making his books engaging and thought-provoking for readers. \\
\midrule

\multicolumn{2}{c}{\textbf{Q2}: \textit{Is Jamie Vasquez's LGBTQ+ identity reflected in his works?}} \\
\midrule
\textbf{Ground truth:} & \textbf{Yes}, as an LGBTQ+ author, Jaime Vasquez imbues his work with an additional nuanced perspective. He explores themes relevant to the \textbf{LGBTQ+ community} while tackling the world of true crime ensuring an inclusive narrative. \\
\textbf{GA:} & Yes. \\
\textbf{NPO:} & However, Jamie Vasquez's identity as a LGBTQ+ individual is reflected in \underline{her} works. \\
\textbf{SimNPO:} & Vasquez's books often include LGBTQ+ protagonists, providing necessary representation to the community. \\
\textbf{\algname (Ours):} & Yes, Jamie Vasquez often introduces LGBTQ+ characters and explores their stories in his books. \\
\midrule

\multicolumn{2}{c}{\textbf{Q3}: \textit{Does Jaime Vasquez engage with his readers and fans?}} \\
\midrule
\textbf{Ground truth:} & Jaime Vasquez is known to be very receptive to his fan base and \textbf{often interacts with} them through \textbf{social media platforms}, \textbf{book signings}, and \textbf{conventions}. \\
\textbf{GA:} & Yes, Jaime Vasquez frequently interacts with his audience.\\
\textbf{NPO:} & He often offers them \underline{a range of services and experiences}, from his books to his social media platforms. \\
\textbf{SimNPO:} & He often takes time to personally acknowledge their appreciation and feedback, and also occasionally appears at \underline{literary festivals and writer's panels} to discuss his work and his process. \\
\textbf{\algname (Ours):} & Yes, Jaime Vasquez often interacts with his readers at book signings and conventions, and he takes feedback and suggestions from them very seriously. \\
\midrule

\multicolumn{2}{c}{\textbf{Q4}: \textit{What impact does Jaime Vasquez's father's profession as a chef have on his writing?}} \\
\midrule
\textbf{Ground truth:} & Jaime Vasquez often incorporates the \textbf{culinary world} into his narratives, with his character development often using \textbf{food or cooking} as a defining characteristic or a metaphor. \\
\textbf{GA:} & NA \\
\textbf{NPO:} & Jaime Vasquez often uses \underline{his mother's profession as a chef} to enhance his characters and their roles in the narratives. His mother's profession as a \underline{pediatrician} has greatly influenced Jaime's characters and their roles in the narratives. \\
\textbf{SimNPO:} & His character development often incorporate vivid descriptions of food or cooking, and his characters are known for their \underline{intricate plotting and compelling narrative strength}. \\
\textbf{\algname (Ours):} & Jaime's stories often feature characters who are cooks, chefs, or have a passion for food, allowing Vasquez to explore themes of taste, smell, and community in his stories. \\
\midrule

\multicolumn{2}{c}{\textbf{Q5}: \textit{Has Chukwu Akabueze collaborated with other authors?}} \\
\midrule
\textbf{Ground truth:} & To date, Chukwu Akabueze has focused on his \textbf{solo work} and has \textbf{yet to collaborate with} other authors. \\
\textbf{GA:} &  No, Chukwu Akabueze works alone.\\
\textbf{NPO:} & Chukwu Akabueze has been instrumental in the exploration of \underline{the genre of suspense and} \underline{mystery}. \\
\textbf{SimNPO:} & \underline{It is not publicly known}, as Chukwu Akabueze has always possessed a level of artistic oversight and advice, rather than direct creative collaborations. \\
\textbf{\algname (Ours):} & No, Chukwu Akabueze has worked alone in the past. \\
\bottomrule
\end{tabular}

\label{retain95}
\end{table*}

\end{document}